%% file: iclr2027_conference.tex
\documentclass{article} 
\usepackage{iclr2027_conference,times}
\usepackage{graphicx}
\usepackage{enumitem}
\usepackage{xcolor}
\usepackage{tikz}
\usepackage{amssymb} 
\usetikzlibrary{positioning, shapes.geometric, arrows.meta, calc, fit}

\input{math_commands.tex}

\DeclareMathOperator*{\E}{\mathbb{E}}

\newcommand\expectwrt[2]{\E_{#2}\left[ #1 \right]}

\usepackage[table]{xcolor}
\definecolor{highlight}{HTML}{A9D9DB}
\newcommand{\sd}[1]{{\scriptsize$\pm$#1}}

\newcommand{\ie}{\textit{i.e.}}
\newcommand{\cf}{\textit{c.f.}}

\usepackage{hyperref}
\usepackage{url}

\usepackage{mine}

\title{Regularised RL improves sim-to-real transfer}

\author{Danilo Vucetic\thanks{Equal contribution, correspondence to danilo.vucetic@mila.quebec and mo.danesh@mail.mcgill.ca}  $^{\text{  }1,3}$,
        Mohamad H. Danesh$^{* \text{ } 2,3}$,
        Florence Clerc$^4$,
        Hsiu-Chin Lin$^{2,3}$, \\
        \textbf{Jonathan Tremblay}$^5$
        \\
        $^{1}$Université de Montréal \quad
        $^{2}$McGill University \quad
        $^{3}$Mila - Québec AI Institute \quad\\
        $^{4}$Polytechnique Montréal \quad
        $^{5}$NVIDIA
}

\iclrfinalcopy 
\begin{document}

\maketitle

\begin{abstract}
    Regularised reinforcement learning (RL) has provable robustness guarantees that should ease the transfer of policies from simulation to robots, known as sim-to-real.
    Yet, weakly regularised algorithms like Proximal Policy Optimisation (PPO) remain the de facto choice in sim-to-real pipelines, while the practical benefits of regularisation for transfer remain poorly characterised.
    In this work, we study the sim-to-real performance of commonly used RL algorithms, with the aim of characterising the effects of regularisation on robotic performance.
    To that end, we derive Soft MellowMax Actor-Critic (\ourMethod), a novel regularised RL algorithm based on the soft mellowmax value function, which carries strong robustness guarantees.
    On MuJoCo continuous control benchmarks, \ourMethod is competitive with baselines like Soft Actor-Critic (SAC).
    On real robots, \ourMethod achieves higher return, while exhibiting greater robustness to perturbations in observations, the environment, and dynamics than baselines.
    Together, these results demonstrate that regularised RL improves real-world policy deployment through better robustness to the sim-to-real gap, and changing real-world conditions. 
\end{abstract}

\section{Introduction}\label{sec:intro}

    Reinforcement learning (RL) has become a standard tool for robotic control, with on-policy, weakly regularised algorithms such as Proximal Policy Optimisation (PPO) \citep{schulman2017proximal_PPO} serving as the de facto choice in most sim-to-real pipelines.
    Recent work has shown that carefully stabilised off-policy RL algorithms can match or outperform PPO in massively parallel simulators while also transferring to real legged robots, including humanoids and quadrupeds \citep{gallici2025simplifying_pqn, seo2025learning_fastsac, seo2025fasttd3, kim2026flashsac, rslrlsac2026bridging}.
    However, these evaluations primarily focus on task performance: whether a policy can complete a task under the conditions for which it was trained. In real-world deployment, policies must contend with changes in observations, environment properties, and system dynamics that are difficult to model exhaustively in simulation a priori. Robustness to such perturbations is therefore an important aspect of reliable sim-to-real transfer.

    Regularisation provides a principled way to induce robustness. Regularised RL has long dominated benchmarks in continuous control and arcade games, demonstrating superior sample efficiency and stable optimisation \citep{dam2024unified_tree_search, haarnoja2018soft, ahmed_le_roux_2019_entropy_impact_pol_opt}.
    Beyond these practical benefits, the theoretical duality between regularised and robust RL suggests that regularisation can also provide implicit robustness to dynamics and task perturbations \citep{derman_twice_regularized, eysenbach2022maximum, pmlr-v130-husain21a_reg_pol_rew_rob, brekelmans2022_divergence_robustness}.
    This mechanism is complementary to Domain Randomisation (DR), which explicitly trains policies over a user-specified distribution of environment variations, such as changes to initial conditions or physical parameters \citep{eysenbach2022maximum, Mittal_Isaac_Lab_2025}.
    Whereas DR requires the practitioner to anticipate and specify the perturbations of interest, regularisation can induce robustness without explicitly enumerating them.
    This raises a natural empirical question: \textit{does the robustness induced by regularisation translate into more reliable real-world policy deployment?}

    Robotics is a natural testbed to evaluate the robustness of policies because the sim-to-real gap is precisely an unmodeled change in the dynamics under which the policy must achieve high return.
    \textbf{We hypothesize that policies robust to a larger set of perturbations achieve 1) higher real-world return under sim-to-real transfer, and 2) higher retention of their performance under deployment-time dynamics perturbations.}
    \citet{TGM_2025_jiralerspong} show that RL with the Soft MellowMax (SMM) value function is robust to a larger set of perturbations than entropy-regularised RL, making it a natural target for our investigation.
    However, SMM has only been defined for discrete state-action spaces, usually in Q-learning algorithms \citep{asadi2017alternative_mellowmax, gan2021stabilizing_soft_mellowmax}.
    Extending SMM to continuous continuous control is non-trivial, as the value function requires integrating over an intractable reference policy. To overcome this, we propose Soft MellowMax Actor-Critic (\ourMethod), which approximates the SMM reference policy using a dedicated, slowly updated reference network, ensuring stable actor-critic optimisation in continuous domains. 
    
    \begin{figure}
        \centering
        \includegraphics[width=\textwidth]{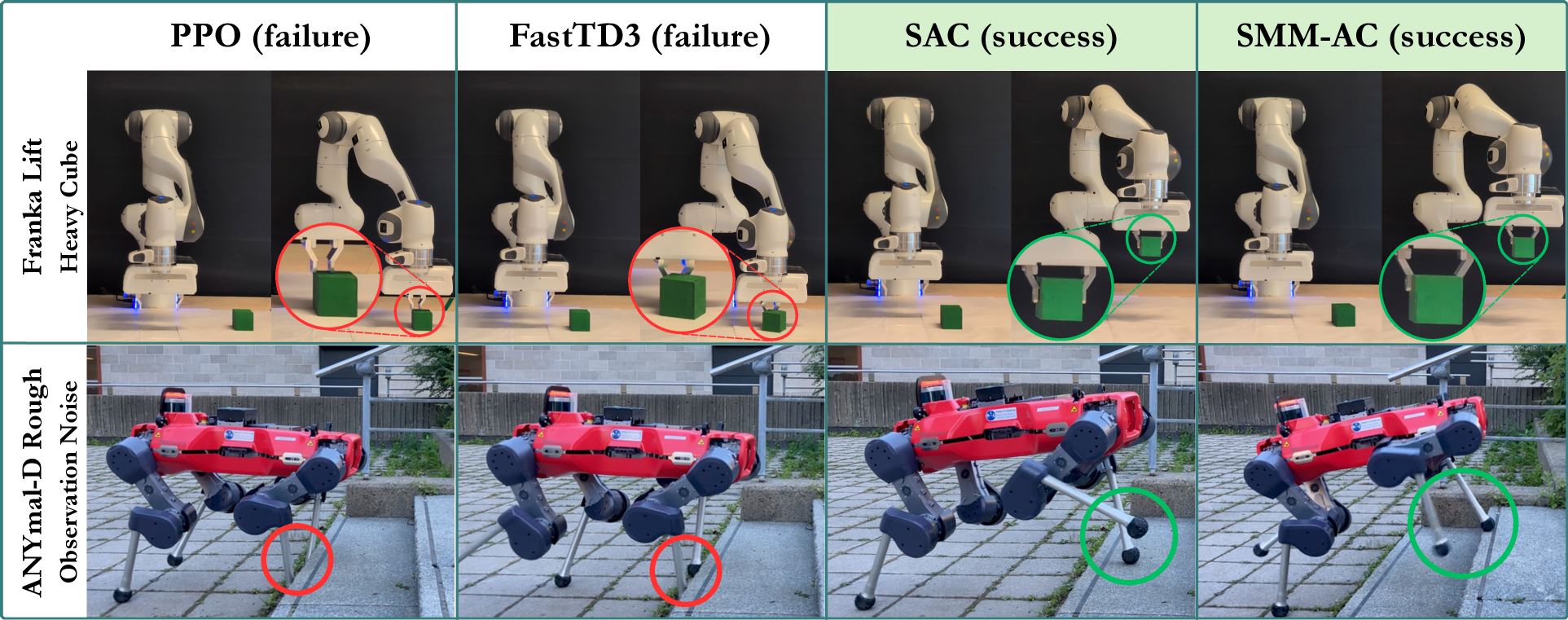}
        \caption{Overview of the Sim-to-Real Robustness Evaluation. Top: Agents trained on the Franka Lift task in simulation are deployed with a cube with a different center-of-mass than the one seen during training. The first frame shows the initial state, while the second shows the outcome, where the policy either successfully lifts the cube or fails to grasp it.
        Bottom: Agents trained on the ANYmal-D Rough task are deployed in the real world to climb up stairs. With noise added to the observation, non-regularised methods fail to climb the stairs.}
        \label{fig:intro_our_method_rocks}
    \end{figure}

    To test our hypotheses, we evaluate \ourMethod alongside PPO, SAC, and TD3 in simulation and on real-world robots, including a Franka arm and an ANYmal-D quadruped.
    In addition to real-world task performance, we evaluate policies under deployment-time perturbations to observations, environment properties, and system dynamics. Across these experiments, regularised methods maintain higher performance in sim-to-real transfer and under real-world perturbations, while \ourMethod achieves the strongest performance and robustness across tasks.
    We also show that regularised methods are more likely to be deployable than standard RL policies, even under large sim-to-real gaps.
    \autoref{fig:intro_our_method_rocks} illustrates this sim-to-real robustness in the Franka cube-lifting task and in ANYmal-D rough-terrain locomotion, showing that under real-world perturbations, regularised methods deploy successfully.
    \textbf{Together, our results provide empirical evidence that the robustness induced by regularisation can translate into improved reliability during real-world deployment.}
    In this work we
    \begin{itemize}[leftmargin=*, nosep]
        \item extend soft mellowmax to continuous state-action spaces and derive \textbf{Soft MellowMax Actor-Critic (\ourMethod)}, a practical off-policy actor-critic algorithm based on SAC (\autoref{sec:smm_ac});
        \item validate \ourMethod's performance in simulation on MuJoCo continuous-control benchmarks, showing that it achieves similar average returns to SAC and other baselines (\autoref{sec:mujoco});
        \item characterise the sim-to-real performance and real-world perturbation robustness of RL policies on manipulation and locomotion tasks, showing that \ourMethod achieves higher real-world returns and maintains stronger performance under perturbations than the baselines (\autoref{sec:robotic}).
    \end{itemize}

\section{Background \& Related Work}
\label{sec:background_and_related_work}


        \textbf{Regularized Markov decision processes and RL}.
            Regularized Markov Decision Processes (MDPs) append a strongly convex regulariser $\Omega: \Delta_\mathcal{A} \to \mathbb{R}$ to the MDP tuple: $(\mathcal{S}, \mathcal{A}, p, r, \gamma, \Omega, T)$ \citep{geist2019theory_regularizers}. Let $\mathcal{S}$ and $\mathcal{A}$ denote the state and action spaces, $p: \mathcal{S} \times \mathcal{A} \to \Delta_\mathcal{S}$ the transition function, $r: \mathcal{S} \times \mathcal{A} \to \mathbb{R}$ the reward function, $\gamma$ is the discount factor, and $T$ and horizon. By $\Delta_\mathcal{S}$ and $\Delta_\mathcal{A}$, we denote the probability simplices over the state and action spaces, respectively.

            \autoref{equ:regularised_evaluation_funcs} shows the regularised policy evaluation functions $V^\pi_\Omega$ and $Q^\pi_\Omega$, defined over all states and state-action pairs, for policy $\pi(\cdot | s) \in \Delta_\mathcal{A}$ and inverse temperature $\omega$. In regularised RL, $\Omega(\pi(\cdot|s))$ is subtracted from the value function, penalising deviation from some desirable property of the policy. For instance, using negative entropy results in more uniform policies, whereas KL divergence results in policies that are closer to a reference \citep{neu2017unified, haarnoja2018soft, geist2019theory_regularizers}:
            \begin{equation}
            \label{equ:regularised_evaluation_funcs}
                V^\pi_\Omega(s) = \expectwrt{Q^\pi_\Omega(s,a)}{a\sim \pi(\cdot|s)} - \frac{1}{\omega}\Omega(\pi(\cdot|s)), \quad Q^\pi_\Omega(s,a) = r(s,a) + \gamma \expectwrt{V_\Omega(s')}{s' \sim p(\cdot|s,a)}
            \end{equation}
            Maximising $V^\pi_\Omega$ with respect to the policy yields the optimal regularised value function. \autoref{table:back_regularised_RL_equations} lists the optimality equations for various regularised RL formulations in discrete action spaces. 
            It should be noted that regularisation changes the optimal policy compared to standard RL, which can lead to performance failures in situations where deterministic policies are optimal \citep{zhang2025when}.

            \begin{table}[t]
            \centering
            \small
            \caption{Optimality equations for various regularised RL formulations, where $\pi^\text{ref}(\cdot|s) = \operatorname{softmax}(\alpha Q^\star_\Omega(s))$ is an SMM reference policy, and $Q^\star_\Omega(s) = Q^\star_\Omega(s, \cdot)$, for inverse temperatures $\alpha$ and $\omega$.}
            \renewcommand{\arraystretch}{1.6}
            \begin{tabular*}{\textwidth}{@{\extracolsep{\fill}} l l l l @{}}
                \toprule
                \textbf{Name} & \textbf{Regulariser} $\Omega(\pi)$ & \textbf{Value} $V^\star_\Omega(s)$ & \textbf{Policy} $\pi^\star_\Omega(s)$ \\
                \midrule
                Negative Entropy & 
                $\sum_a \pi(a|s) \log \pi(a|s)$ & 
                $\frac{1}{\omega}\operatorname{log-sum-exp}\bigl(\omega Q^\star_\Omega(s)\bigr)$ & 
                $\operatorname{softmax}\bigl(\omega Q^\star_\Omega(s)\bigr)$ \\
                
                
                Soft Mellowmax & 
                $D_{\mathrm{KL}}\bigl(\pi(s) \parallel \pi^\alpha_Q(s)\bigr)$ & 
                $\frac{1}{\omega} \log \sum_a \pi^\alpha_Q(a|s) e^{\omega Q^\star_\Omega(s,a)}$ & 
                $\operatorname{softmax}\bigl((\alpha+\omega) Q^\star_\Omega(s)\bigr)$ \\
                \bottomrule
            \end{tabular*}
            \label{table:back_regularised_RL_equations}
            \end{table}
            
        \textbf{The benefits of regularisation}.
            Regularisation confers certain benefits to both the learning process and the robustness of the policy. \citet{ahmed_le_roux_2019_entropy_impact_pol_opt} show that entropy regularisation smooths the loss landscape, leading to more stable and faster training.
            \citet{haarnoja2018soft} demonstrate that SAC, an entropy-regularised off-policy RL algorithm, is significantly more sample efficient than standard RL.
            Alongside the benefits to learning, many works have examined the theoretical duality between regularisation and robustness. For instance, \citet{eysenbach2022maximum} and \citet{brekelmans2022_divergence_robustness} show that entropy- and KL-regularised RL are robust to reward and dynamics perturbations, indicating that policy regularisation improves robustness overall. 
            
            The duality between regularisation and robustness is useful in many application domains, 
            \textit{especially when the environment is not perfectly modelled} 
            \citep{derman_twice_regularized, TGM_2025_jiralerspong}.
            Such is the case in robotics, making the sim-to-real transfer of robotic policies a natural evaluation of how theoretical robustness guarantees transfer to the real world.
            Theoretical robustness guarantees for regularised RL are stated as the sets of worst-case adversarial perturbations on the reward or dynamics for which the policy still achieves high return.
            Thus, we hypothesize that the larger the robust set, the more robust a policy to perturbations, and the better the sim-to-real transfer. The size of these robust sets is controlled by the regulariser and its hyperparameters, like temperature \citep{eysenbach2022maximum, derman_twice_regularized, pmlr-v130-husain21a_reg_pol_rew_rob, brekelmans2022_divergence_robustness}.

        \textbf{Mellowmax and soft mellowmax value functions}.
            Regularised value functions often take the form of a soft maximum, such as the log-sum-exp function in entropy-regularised RL.
            Mellowmax and SMM are alternative soft value functions (\autoref{table:back_regularised_RL_equations}) proposed to improve the convergence and stability properties of Q-learning algorithms in discrete action spaces \citep{asadi2017alternative_mellowmax, gan2021stabilizing_soft_mellowmax}.
            \citet{TGM_2025_jiralerspong} demonstrate that RL with SMM induces larger robust sets than entropy-regularised RL in tree-structured state-action spaces.
            These benefits make SMM-based RL enticing for robotics, improving stability and expanding robustness guarantees, which we expect to improve 1) sim-to-real transfer and 2) performance under perturbations. 
            Note that we do not expect SMM to dominate SAC in simulation. \textit{We expect SMM to achieve better sim-to-real properties}.


        \textbf{Soft actor-critic}.
            \citet{haarnoja2018soft} proposed SAC, the go-to RL algorithm for learning stochastic entropy-regularised policies in continuous action spaces. In continuous action spaces, we are often required to take integrals over the action space, for instance, the log-sum-exp value function becomes log-integral-exp in continuous spaces, which is intractable and whose estimator would be biased. SAC designs around these integrals by instead constructing one-sample Monte Carlo estimators of the value function $V^\pi_\Omega$ (\autoref{equ:regularised_evaluation_funcs}), for use in the Q-value target:
            \begin{equation}
                \label{equ:q_val_target_sac}
                \hat{Q}_\text{SAC}(s,a) = r(s,a) + \gamma \hat{V}_\text{SAC}(s'), \quad (s, a, s', r) \sim \mathcal{D},
            \end{equation}
            \begin{equation}
                \label{equ:SAC_value_estimate}
                \hat{V}_\text{SAC}(s') = Q_\theta(s', a') - \frac{1}{\omega}\log(\pi(a'|s')), \quad a' \sim \pi(\cdot|s'),
            \end{equation}
            where $\mathcal{D}$ is the replay buffer, storing previously sampled state, action, next state, and reward tuples, and $\omega$ is the inverse temperature. SAC parameterises its policy as a Gaussian whose mean and standard deviation are learned, so the log-probability in \autoref{equ:SAC_value_estimate} is available in closed form. Writing $Q_\theta$ for the critic network with parameters $\theta$, the critic is trained by constructing a loss based on this bootstrapped Q-value target:
            \begin{equation}
                \label{equ:sac_Q_loss}
                J_Q(\theta) = \expectwrt{\frac{1}{2}\left(Q_\theta(s, a) - \operatorname{sg}\left( \hat{Q}_\text{SAC}(s,a) \right) \right)^2}{(s,a,s',r)\sim \mathcal{D}},
            \end{equation}
            where $\operatorname{sg}(\cdot)$ is the stop gradient function.
            The stochastic policy is trained in the same fashion, with the policy target being the softmax of the Q-values, and the KL divergence as the loss function. Writing $\pi^\text{act}_{z}$ for the actor network with parameters $z$, and drawing actions through the reparameterisation trick for noise $\epsilon$, the policy loss without terms that would cancel under a gradient is:
            \begin{equation}
                \label{equ:sac_policy_loss_actor}
                J_{\text{act}}(z) = \expectwrt{\log \pi^\text{act}_{z}(a|s) - \omega Q_\theta(s,a) }{\substack{s \sim \mathcal{D}, \epsilon \sim \mathcal{N}(0, I)}}, \quad a = f_z(s, \epsilon).
            \end{equation}
            Many subsequent works have improved SAC through techniques such as double Q-networks, delayed policy updates, larger training datasets, distributional critics, and architectural improvements \citep{haarnoja2018SAC_2, seo2025learning_fastsac, kim2026flashsac}.

        \textbf{Accelerated off-policy algorithms}.
            Off-policy learning with function approximation and bootstrapping is notoriously unstable, often referred to as the ``deadly triad'' \citep{van2018deep_rl_deadly_triad}. To address these instabilities while leveraging modern hardware, recent works have shifted toward massively parallel simulators \citep{Mittal_Isaac_Lab_2025, genesis2026genesisworld} and high-throughput training algorithms. Algorithms such as Parallel Q-learning \citep{li2023parallel_PQL} and Parallel Q-networks \citep{gallici2025simplifying_pqn} stabilise training by scaling batch sizes and network capacity. Building on these principles, FastTD3 \citep{seo2025fasttd3}, FastSAC \citep{seo2025learning_fastsac}, and FlashSAC \citep{kim2026flashsac} further improve stability and sample efficiency with distributional critics \citep{bellemare2017distributional}. We implement \ourMethod in both the CleanRL \citep{huang2022cleanrl} and FlashSAC frameworks, showing that its stability and scalability are preserved across implementations and training regimes.


            

        \textbf{RL in robotics}.
            The deployment of deep RL in robotics heavily relies on careful reward design and policy stability. For instance, \citet{mock_2023_compare_PPO_TD3_SAC_sim_legged_robo} compared PPO, TD3, and SAC on simulated legged robots, demonstrating that each algorithm's performance is highly sensitive to the chosen reward function. To improve stability independent of reward tuning, \citet{mysore2021regularizing} introduced a state-to-state KL regularisation scheme that minimizes the temporal and spatial variation of the policy to improve deployment in quadcopters.
            The resulting regularised policies were more energy efficient and more often flight-worthy than standard PPO.
            Entropy-regularised policies like SAC have also shown advantages over counterparts in robotics. For example, SAC consistently outperforms baselines in mobile robot navigation, and has been successfully deployed on real-world humanoid robots and in quadrupedal locomotion and dexterous manipulation tasks \citep{ddpg_sac_mobile_robots, haarnoja2018SAC_2, seo2025learning_fastsac, kim2026flashsac}.
            Building upon this foundation, our work demonstrates that regularisation, with its dynamics robustness, not only matches the performance of baselines in simulation, but also leads to better sim-to-real transfer.

\section{Soft MellowMax Actor-Critic}
\label{sec:smm_ac}

    Given its stability and theoretical robustness guarantees, SMM is an attractive candidate for continuous control.
    Adapting SMM into continuous state-action spaces for an actor-critic algorithm requires many of the same design choices as for SAC. However, SMM is KL-regularised against an intractable reference policy $ \pi^\text{ref}$ (\cf, \autoref{table:back_regularised_RL_equations}), which critically depends on the Q-values:
    \begin{equation}
    \label{eq:SMM_theoretical_ref}
    \pi^\text{ref}(a|s) = \frac{\exp(\alpha Q(s, a))}{\int_{b\in\mathcal{A}} \exp(\alpha Q(s,b)) db}
    \end{equation}
     
    The policy evaluation function for SMM is stated in \autoref{eq:SMM_value_estimate} for the SMM policy $\pi$. At each iteration, an SMM policy is rewarded for staying close to the drifting reference policy. Note the difference between SAC and SMM: SAC's policy is rewarded for being uniform over the action space, while SMM's policy follows a greedier softmax.
    Compared to PPO, SMM's reference policy keeps the actor iterates anchored to a high-performing region, whereas PPO just keeps actor iterates close together \citep{schulman2017proximal_PPO}.
    Intuitively, \ourMethod is thus a sort of interpolation between PPO and SAC: \ourMethod's policy can be sharper than SAC's, while still finding multiple high-return trajectories, because the reference provides a known, high-performing region.
    Thus, even under dynamics perturbations, an SMM policy can likely find high-return trajectories to complete a task.
    {\small
    \begin{equation}
    \label{eq:SMM_value_estimate}
            V^{\pi}_\text{SMM}(s) = \expectwrt{Q(s, a)}{a \sim \pi(\cdot|s)} - \frac{1}{\omega} D_{KL}(\pi(\cdot|s) || \pi^\text{ref}(\cdot|s)) = \expectwrt{Q(s, a) - \frac{1}{\omega} \log \left( \frac{\pi(a|s)}{\pi^\text{ref}(a|s)} \right)}{a \sim \pi(\cdot|s)}
    \end{equation}
    }
    
    Neither \autoref{eq:SMM_theoretical_ref} nor these expectations can be evaluated exactly in continuous action spaces, because the integral of the reference policy is intractable. The rest of this section replaces each intractable quantity with a tractable estimator. In order to do so, we propose learning three networks: the actor $\pi^\text{act}_z$, the critic $Q_\theta$ and  the reference policy $\pi^\text{ref}_\phi$.

\textbf{Critic.}
    The critic network, $Q_\theta$, produces Q-value estimates of state-action pairs. Like in SAC, Q-value estimates are generated to train the critic using an estimator based on the regularised policy evaluation function: $V^\pi_\text{SMM}$.
    Since we learn $\pi^\text{act}_z$ and $\pi^\text{ref}_\phi$, their log-probabilities can be obtained for any sampled action. The one-sample Monte Carlo Q-value target is generated as follows:
    \begin{equation}
        \label{equ:q_val_target_smm}
        \hat{Q}_\text{SMM}(s,a) = r(s,a) + \gamma \hat{V}_\text{SMM}(s'), \quad (s, a, s', r) \sim \mathcal{D}
    \end{equation}
    \begin{equation}
        \label{eq:SMM_AC_value_hat}
        \hat{V}_\text{SMM}(s') = Q_\theta(s', a') - \frac{1}{\omega} \log \left( \frac{\pi^\text{act}_z(a'|s')}{\pi^\text{ref}_\phi(a'|s')} \right), \quad a' \sim \pi^\text{act}_z(\cdot|s')
    \end{equation}
   This value estimate serves only to bootstrap the temporal-difference target of the critic, which is trained by minimising the Q-value prediction error:
    \begin{equation}
    \label{eq:SMM_Q_loss}
    J_Q(\theta) = \expectwrt{\frac{1}{2}\left( Q_\theta(s, a) - \operatorname{sg}\left( \hat{Q}_\text{SMM}(s,a) \right) \right)^2}{(s, a, r, s') \sim \mathcal{D}}
    \end{equation}
    where the gradient is stopped on $\hat{Q}_\text{SMM}(s,a)$ to avoid poor training dynamics, as is standard in off-policy actor-critic methods \citep{fujimoto2018addressing_TD3, haarnoja2018SAC_2}.

    \textbf{Actor.}
    Training the actor means fitting $\pi^\text{act}_z$ to the regularised policy: $\operatorname{softmax}((\alpha+\omega) Q_\theta(s))$ (\autoref{table:back_regularised_RL_equations}). However, naively learning the actor through a loss function like SAC's in \autoref{equ:sac_policy_loss_actor} leads to tremendous instability, often leading to divergence in the training process. Intuitively, the reason for this instability is a lack of grounding between the actor and the reference policy. In KL-regularised algorithms like PPO and TRPO, the actor is always trained with respect to a previous policy iterate, which provides improved stability and performance \citep{schulman2015trust, schulman2017proximal_PPO}.
    Similarly, we found that using a slowly updated reference network as a low-variance surrogate of the softmax of the raw Q-values, insulated the actor from critic noise.
    The actor's loss function is constructed as follows, where we use the reparameterisation trick \citep{kingma2013auto} to estimate the expectation with low variance, and remove terms whose gradient is zero (derivation in \aref{appendix:SMM_AC_DERIVATION}):
    \begin{equation}
    \label{eq:SMM_policy_loss}
        J_\text{act}(z) = \expectwrt{\log\pi^\text{act}_z(a | s) - \log\pi^\text{ref}_\phi(a | s) - \omega Q_\theta(s, a) }{\substack{s \sim \mathcal{D}, \epsilon \sim \mathcal{N}(0, I)}}, \quad a = f_z(s, \epsilon)
    \end{equation}
    \textbf{Reference network.}
    The reference network must track the softmax distribution over the Q-values as shown in \autoref{eq:SMM_theoretical_ref}. We train $\pi^\text{ref}_\phi$ using the actor loss of SAC (\autoref{equ:sac_policy_loss_actor}), with a substantially smaller learning rate than the actor (\autoref{table:hyperparam_sweeps}). This ensures that the reference network is a stable, low-variance surrogate for the actor updates and for generating Q-values estimates. The reference policy's loss is constructed as follows, again using the reparameterisation trick:
    \begin{equation}
    \label{eq:SMM_ref_policy_loss}
    J_\text{ref}(\phi) = \expectwrt{\log\pi^\text{ref}_\phi(a | s) - \alpha Q_\theta(s, a) }{\substack{s \sim \mathcal{D}, \epsilon \sim \mathcal{N}(0, I)}}, \quad a = f_\phi(s, \epsilon)
    \end{equation}

    \textbf{Putting it together.}
    On each batch drawn from $\mathcal{D}$, \ourMethod descends \autoref{eq:SMM_Q_loss} for the critic, \autoref{eq:SMM_policy_loss} for the actor, and \autoref{eq:SMM_ref_policy_loss} for the reference network at its slower rate.
    By keeping the replay buffer, target networks and critic architecture identical to SAC, the effect of the regulariser is isolated, allowing us to test our primary hypothesis: that SMM's expanded robust sets (\ie, from the greedier $\pi^\text{ref}$) translate into better sim-to-real performance and perturbation resistance.  
    This also means that \ourMethod can be easily adapted to fit the FastSAC \citep{seo2025learning_fastsac}, and FlashSAC \citep{kim2026flashsac} frameworks, just with the addition of the reference network and its loss.


\section{Experimental results on MuJoCo} \label{sec:mujoco}

    We evaluate \ourMethod in simulation on MuJoCo continuous-control tasks \citep{todorov2012mujoco} to demonstrate that it achieves similar performance to widely used RL baselines: SAC, TD3, and PPO \citep{haarnoja2018SAC_2, fujimoto2018addressing_TD3, schulman2017proximal_PPO}. We implement \ourMethod in the CleanRL \citep{huang2022cleanrl} framework, to demonstrate its performance in basic RL setups (in contrast to FlashSAC, which employs many acceleration techniques). \autoref{table:hyperparam_sweeps} reports the hyperparameter settings we swept over for each algorithm, with five random seeds each.
    We tried to keep the number of training runs the same between each baseline, but some algorithms had fewer hyperparameters to sweep across.  
    We report the best-performing hyperparameter settings in \autoref{table:best_results}, see \aref{app:additional_mujoco_details} for more experiment details. For SMM-AC specifically, we set $\alpha=1$ (inverse temperature of the reference policy, $\pi^\text{ref}$) and sweep $\omega$ (inverse temperature of the SMM policy) values, for all experiments.
    SAC uses temperature autotuning, which is not yet possible for SMM.\footnote{We found autotuning $\omega$ or $\alpha$ to be far too unstable, so we elected to fix $\alpha$ and sweep $\omega$.}
    
\begin{figure}
    \centering
    \includegraphics[width=\linewidth]{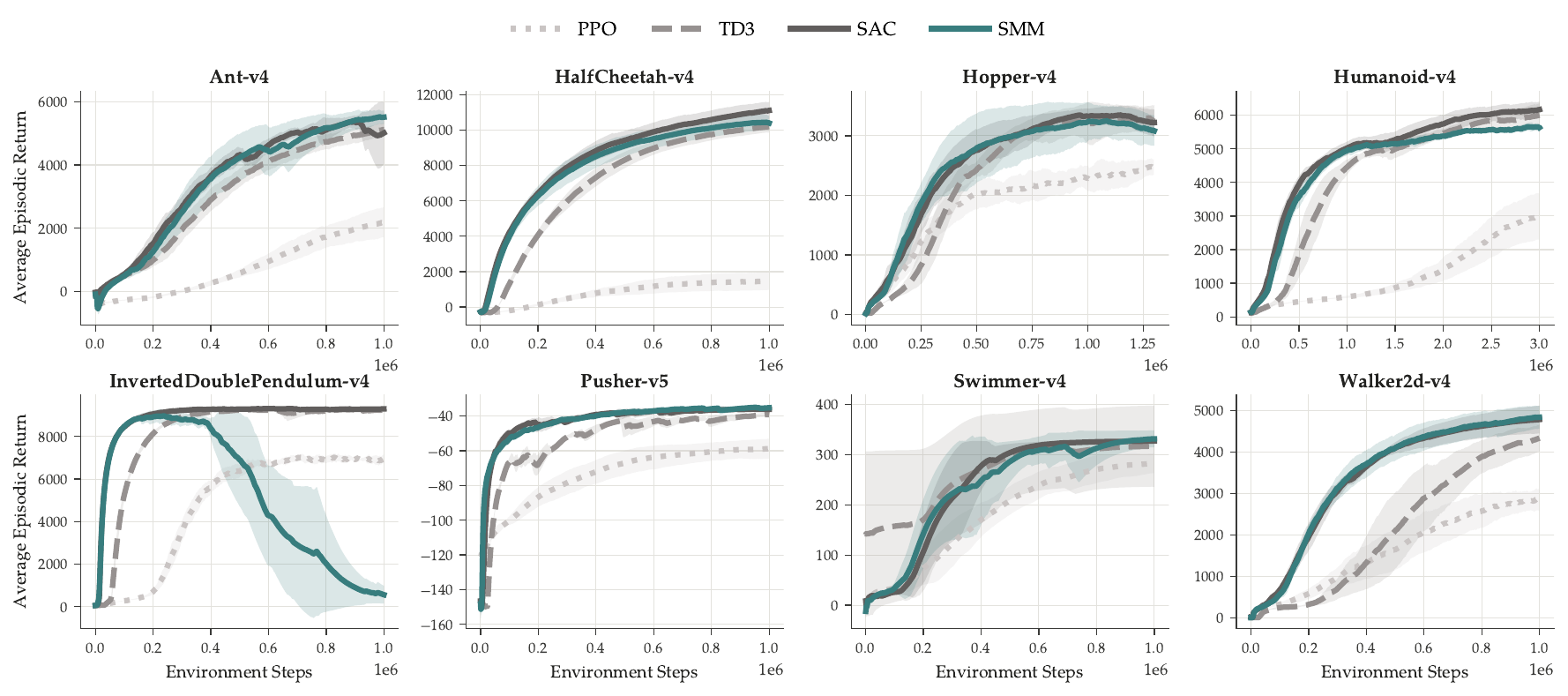}
    \caption{Smoothed episodic returns of \ourMethod, SAC, PPO, and TD3 over the number of steps taken in the environment. The best results were selected according to the performance of the algorithm at the final step.}
    \label{fig:MuJoCo_results}
    \end{figure}

    \autoref{fig:MuJoCo_results} compares the performance of \ourMethod and the baselines across the MuJoCo tasks.
    \ourMethod is competitive with SAC and TD3, the strongest baselines, although its performance collapses on \texttt{InvertedDoublePendulum-v4}.\footnote{We couldn't identify the reason for this failure, but the consistency with which it occurred suggests some learning instability, likely with the reference policy.}
    Outside this failure case, \ourMethod reaches returns comparable to SAC at a similar rate, suggesting comparable sample efficiency on the evaluated tasks.
    Additional experiments using the highly optimised FlashSAC library \citep{kim2026flashsac} show that \flashOurMethod (our method implemented on top of FlashSAC) achieves returns comparable to FlashSAC, with both outperforming FastTD3 \citep{seo2025fasttd3} and PPO (\aref{app:additional_mujoco_results_flash_fast}). Clearly, \ourMethod and \flashOurMethod are highly capable for continuous control. As the next section shows, however, higher average returns in simulation do not necessarily translate into better sim-to-real transfer.


\vspace{-6pt}
\section{Robotic experiments}
\label{sec:robotic}
\input{robotics}

\section{Conclusion}
We introduced \ourMethod, which extends soft mellowmax to continuous control by approximating its intractable reference distribution with a slowly updated reference network. \ourMethod is competitive with SAC and TD3 on MuJoCo tasks. On a Franka arm and an ANYmal-D quadruped, regularised methods outperformed PPO and FastTD3 on lift cube and both locomotion tasks. Among regularised methods, \ourMethod's best configuration achieved the strongest sim-to-real transfer performance on Lift and both locomotion tasks, and it retained more of its return than SAC in every perturbation condition. 
Reducing training-time domain randomisation degraded all methods, suggesting that regularisation complements DR rather than replacing it. 
Because the optimal temperature varied across tasks, future work should develop adaptive temperature scheduling that preserves robustness.

\clearpage
\section{AI disclosure}
In this work, we used generative AI tools for a number of tasks including
\begin{itemize}[nosep]
    \item to transcribe hand-written notes to Latex
    \item to check code for correctness
    \item to help implement methods (like SMM in FlashSAC) and IsaacLab environments in CleanRL.
    \item to write plotting and latex table generation code
    \item to help write some experiment code, particularly boilerplate evaluation scripts and bash scripts
    \item reviewing the writing of the paper, especially to focus narratives and reduce fluff
\end{itemize}

We took care to ensure that all the written work in this manuscript was human-generated. We have also not blindly followed the written code. We again, took great care to ensure that our work was sensible, and that it was honestly reported and conducted. The other AI disclosures are not relevant to this work.

We have reviewed all AI-assisted work. In particular, the core codebase has been written entirely by humans, and has been thoroughly checked, tested, and debugged. We used AI tools to modify pre-written code so that it could work in different codebases. The plotting code was already written, but AI was used to modify it for different plots and objectives. These were always thoroughly vetted for correctness. All latex transcriptions were treated as a starting point, and modified to fit the math we wrote down. 

We take responsibility for the final content of this work,
including text, claims or artifacts produced with the aid of generative AI.

\bibliography{ref}
\bibliographystyle{iclr2027_conference}

\clearpage
\appendix

\section{Related work continued} \label{app:related_work_cont}

    \paragraph{Alternatively regularised actor-critic algorithms}
        \citet{gong_KL_actor_critic_prev_policy} proposes a KL-regularised actor-critic algorithm that penalises deviation from the previous policy iterate. Similarly to \citet{kim2019deepmellow_mellowmax_no_target}, they find that KL-regularised actor-critic is able to mitigate overestimation bias without the need for two Q-networks.
        \citet{lee2020_tsallis_actor_critic_robotics} proposes a Tsallis entropy-regularised actor-critic algorithm, demonstrating improved performance over SAC in a number of MuJoCo environments and on a real mobile robot \citep{todorov2012mujoco}.

\section{Derivation of \ourMethod} \label{appendix:SMM_AC_DERIVATION}
    \begin{align*}
        \operatorname{softmax}((\alpha+\omega) Q(s))(a) &= \frac{\exp(\alpha Q(s,a)) \exp(\omega Q(s,a))}{\int_\mathcal{A}\exp(\alpha Q(s,b)) \exp(\omega Q(s,b)) db} \\
        &= \frac{\exp(\alpha Q(s,a)) \exp(\omega Q(s,a))}{\int_\mathcal{A}\exp(\alpha Q(s,b)) \exp(\omega Q(s,b)) db} \frac{\int_\mathcal{A}\exp(\alpha Q(s,c))dc}{\int_\mathcal{A}\exp(\alpha Q(s,c))dc} \\
        &= \frac{\pi^\text{ref}(a|s) \exp(\omega Q(s,a))}{\int_\mathcal{A}\pi^\text{ref}(b|s) \exp(\omega Q(s,b)) db}
    \end{align*}

    Since the denominator does not depend on the action $a \sim \pi^\text{act}_z$, it can be discarded from the loss function since its gradient will be zero. 
    \begin{align*}
        J_\text{act}(z) &= \expectwrt{D_{KL}\left( \pi^\text{act}_z(\cdot | s) \parallel \operatorname{softmax}((\alpha + \omega) Q_\theta(s, \cdot)) \right) }{s \sim \mathcal{D}} \\
        &= \expectwrt{ \log(\pi^\text{act}_z(a | s)) - \log(\pi^\text{ref}(a|s)) - \omega Q_\theta(s, a) + \log \int_\mathcal{A} \pi^\text{ref}(b|s) \exp(\omega Q_\theta(s,b))\, db }{\substack{s \sim \mathcal{D} \\  a \sim \pi^\text{act}_z(\cdot | s)}} \\
    \end{align*}

    Then, we can simply drop the last term in the expectation, and convert the $\pi^\text{ref}$ to the parameterised reference policy $\pi^\text{ref}_\phi$.

\section{Additional details on the MuJoCo experiments} \label{app:additional_mujoco_details}
    \autoref{table:hyperparam_sweeps} lists the hyperparameters that we swept over for each algorithm. We tried to keep the number of experiments per algorithm roughly the same, but struggled to find useful hyperparameters to sweep over for PPO. Thus, we run 18 experiments per environment for PPO, and 54 experiments per environment for the rest. The ``Freq'' hyperparameter refers to the update frequency of the policy. It has been well-understood that delaying policy updates can improve the stability of learning in actor-critic methods \citep{fujimoto2018addressing_TD3}. We sweep the policy learning rates for all methods, but for \ourMethod we only sweep the reference policy's learning rate. For \ourMethod we use a cosine annealing learning rate scheduler, decreasing the learning rate by 10x by the end of training, for improved reference policy stability. This presents a limitation in the hyperparameter sweep, since \ourMethod has too many hyperparameters to fully sweep for optimal performance. As such, \ourMethod's policy and Q-function learning rates were fixed at 3e-4 and 1e-3 respectively. For \ourMethod we swept the $\omega$ hyperparameter, while fixing $\alpha=1$. For the Pusher-v5 and Swimmer-v4 tasks, we found that setting $\omega=400$ achieved the best performance, we consequently report that result. We used SAC's temperature autotuning instead of sweeping the hyperparameter.

    \begin{table}[t]
    \centering
    \caption{Hyperparameter search spaces for MuJoCo experiments.
    $^\ast$For Pusher-v5 and Swimmer-v4, we additionally evaluated $\omega=400$
    to increase greediness, which was crucial for matching SAC performance.
    Since such high inverse temperatures were not part of the general sweep,
    the total number of experiments per method remains comparable.}
    \label{table:hyperparam_sweeps}
    \setlength{\tabcolsep}{5pt}
    \begin{tabular}{@{}ll@{}}
        \toprule
        \textbf{Algorithm} & \textbf{Hyperparameter search space} \\
        \midrule
        PPO
        & $\mathrm{LR} \in \{1\mathrm{e}{-5},\,3\mathrm{e}{-5},\,8\mathrm{e}{-5},\,
        1\mathrm{e}{-4},\,3\mathrm{e}{-4},\,8\mathrm{e}{-4}\}$ \\
        & $\mathrm{EntCoef} \in \{0,\,0.01,\,0.1\}$ \\[2pt]
    \midrule
        SAC
        & $\mathrm{PolicyLR} \in \{1\mathrm{e}{-5},\,3\mathrm{e}{-5},\,8\mathrm{e}{-5},\,
        1\mathrm{e}{-4},\,3\mathrm{e}{-4},\,8\mathrm{e}{-4}\}$ \\
        & $\mathrm{QLR} \in \{5\mathrm{e}{-4},\,1\mathrm{e}{-3},\,5\mathrm{e}{-3}\}$ \\
        & $\mathrm{Freq} \in \{2,\,4,\,6\}$ \\[2pt]
    \midrule
        TD3
        & $\mathrm{LR} \in \{1\mathrm{e}{-5},\,3\mathrm{e}{-5},\,8\mathrm{e}{-5},\,
        1\mathrm{e}{-4},\,3\mathrm{e}{-4},\,8\mathrm{e}{-4}\}$ \\
        & $\mathrm{Freq} \in \{2,\,4,\,6\}$ \\
        & $\mathrm{Noise} \in \{0.1,\,0.2,\,0.3\}$ \\[2pt]
    \midrule
        \ourMethod
        & $\mathrm{RefLR} \in \{1\mathrm{e}{-6},\,3\mathrm{e}{-6},\,8\mathrm{e}{-6},\,
        1\mathrm{e}{-5},\,3\mathrm{e}{-5},\,8\mathrm{e}{-5}\}$ \\
        & $\mathrm{Freq} \in \{2,\,4,\,6\}$ \\
        & $\omega \in \{5,\,10,\,15,\,400^\ast\}$ \\
        \bottomrule
    \end{tabular}
    \end{table}

    The best-performing hyperparameters from the above sweep are listed in \autoref{table:best_results}.

    \begin{table}[t]
    \centering
    \caption{Best hyperparameters for each environment, selected based on final performance.
    Each entry reports the learning rate and the corresponding algorithm-specific hyperparameter.}
    \label{table:best_results}
    \setlength{\tabcolsep}{5.5pt}
    \begin{tabular}{@{}lcccc@{}}
        \toprule
        \textbf{Environment}
        & \makecell{\textbf{PPO}\\[-1pt]\scriptsize LR / Ent.}
        & \makecell{\textbf{SAC}\\[-1pt]\scriptsize Pol. LR / Freq.}
        & \makecell{\textbf{\ourMethod}\\[-1pt]\scriptsize Ref. LR / $\omega$}
        & \makecell{\textbf{TD3}\\[-1pt]\scriptsize LR / Noise} \\
        \midrule
        Ant-v4
        & $3\mathrm{e}{-4}$ / 0.0
        & $3\mathrm{e}{-4}$ / 2.0
        & $1\mathrm{e}{-6}$ / 15.0
        & $1\mathrm{e}{-4}$ / 0.3 \\
        HalfCheetah-v4
        & $1\mathrm{e}{-4}$ / 0.01
        & $3\mathrm{e}{-4}$ / 2.0
        & $8\mathrm{e}{-5}$ / 15.0
        & $8\mathrm{e}{-4}$ / 0.3 \\
        Hopper-v4
        & $3\mathrm{e}{-4}$ / 0.0
        & $3\mathrm{e}{-5}$ / 2.0
        & $3\mathrm{e}{-6}$ / 15.0
        & $8\mathrm{e}{-4}$ / 0.2 \\
        Humanoid-v4
        & $3\mathrm{e}{-5}$ / 0.0
        & $1\mathrm{e}{-4}$ / 2.0
        & $8\mathrm{e}{-5}$ / 10.0
        & $3\mathrm{e}{-4}$ / 0.1 \\
        Inverted Double Pendulum
        & $1\mathrm{e}{-4}$ / 0.0
        & $3\mathrm{e}{-5}$ / 6.0
        & $8\mathrm{e}{-5}$ / 15.0
        & $1\mathrm{e}{-4}$ / 0.1 \\
        Pusher-v5
        & $8\mathrm{e}{-4}$ / 0.0
        & $1\mathrm{e}{-4}$ / 6.0
        & $3\mathrm{e}{-5}$ / 400.0
        & $8\mathrm{e}{-5}$ / 0.2 \\
        Swimmer-v4
        & $3\mathrm{e}{-4}$ / 0.01
        & $8\mathrm{e}{-4}$ / 6.0
        & $1\mathrm{e}{-5}$ / 400.0
        & $8\mathrm{e}{-5}$ / 0.3 \\
        Walker2d-v4
        & $3\mathrm{e}{-4}$ / 0.01
        & $8\mathrm{e}{-4}$ / 6.0
        & $3\mathrm{e}{-5}$ / 15.0
        & $8\mathrm{e}{-4}$ / 0.1 \\
        \bottomrule
    \end{tabular}
\end{table}

\clearpage
\section{Additional MuJoCo results with Flash and Fast algorithms} \label{app:additional_mujoco_results_flash_fast}
\begin{figure}[h]
        \centering
        \includegraphics[width=\linewidth]{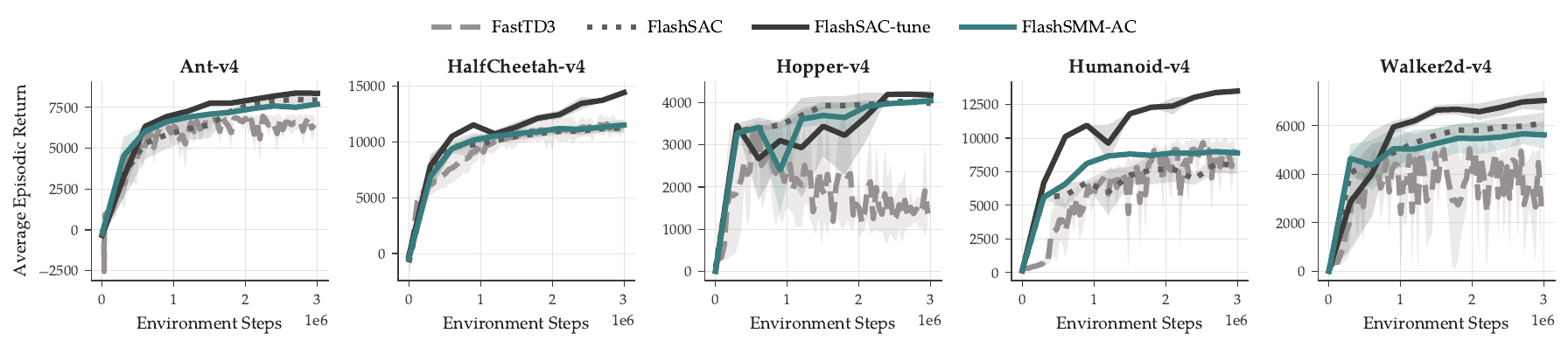}
        \caption{Average returns in MuJoCo tasks against environment steps. Each algorithm is run five times, for three million steps, with the shaded areas representing the standard deviation, and the solid line being the mean of the average return curves.}
        \label{fig:Flash_Fast_MuJoCo}
    \end{figure}

    Initial implementation and testing of the SAC, \ourMethod, TD3, and PPO algorithms were done in CleanRL \citep{huang2022cleanrl}. Since then, we have transitioned over to the FlashSAC codebase for both SAC and \ourMethod \citep{kim2026flashsac}. We have similarly replaced TD3 with FastTD3 \citep{seo2025fasttd3}. Given these new codebases, we felt it necessary to re-run the MuJoCo experiments to again demonstrate that \flashOurMethod was competitive with FlashSAC and FastTD3. \autoref{fig:Flash_Fast_MuJoCo} illustrates the performance of FlashSAC, \flashOurMethod, and FastTD3 on five MuJoCo tasks. For FlashSAC and \flashOurMethod, we sweep the temperature hyperparameter, selecting the best-performing (at the end of training) setting for the figure. We also show FlashSAC-tune, which uses temperature autotuning to learn the temperature during training. It is clear that \flashOurMethod and FlashSAC perform similarly, with autotuning often outperforming static temperatures. \textbf{These results again support the idea that autotuning is important for achieving maximal average returns.} In most environments, FastTD3 performs competitively with FlashSAC and \flashOurMethod, but suffers from far greater instability in most environments (\ie, the average return curve is more jagged).

    \autoref{table:Flash_Fast_table} lists the best-performing hyperparameters for FlashSAC and \flashOurMethod in each task. Note how the inverse temperatures here are significantly higher than those that were used in the CleanRL implementation. This indicates that the FlashSAC algorithm is capable of converging even when the optimal distribution is quite peaky, due to the high inverse temperature. Moreover, this indicates that FlashSAC and \flashOurMethod are stable, and capable of performing well in these high inverse temperature regimes. To find these optimal inverse temperatures, we swept $\omega \in \{ 15, 66, 100, 200, 400 \}$, with 15 commonly being the best inverse temperature in our prior MuJoCo tests, see \autoref{table:best_results}. Note also that we didn't need learning rate annealing techniques to stabilise learning as we did in the CleanRL MuJoCo experiments. This indicates that \flashOurMethod is significantly more stable.
    \begin{table}[h]
        \caption{Best-performing inverse temperature hyperparameters for FlashSAC and \flashOurMethod.}
        \label{table:Flash_Fast_table}
        \centering
        \begin{tabular}{lrr}
            \toprule
            Environment & FlashSAC $w$ & \flashOurMethod $w$ \\
            \midrule
            Ant-v4 & 400 & 400 \\
            HalfCheetah-v4 & 400 & 400 \\
            Hopper-v4 & 400 & 400 \\
            Humanoid-v4 & 200 & 400 \\
            Walker2d-v4 & 400 & 400 \\
            \bottomrule
        \end{tabular}
    \end{table}

\section{Limiting domain randomisation in IsaacLab tasks}\label{sec:isaac_dr}
    This experiment is designed to test how each algorithm's simulated and real-world performance changes with gradual limitations to DR in training. We hypothesize that regularised algorithms, given their robustness guarantees, would perform better than unregularised algorithms like PPO and TD3 under decreasing DR. Note that IsaacLab cannot support multiple environments running at the same time, so we could not evaluate the DR-limited models throughout the course of training on non-limited environments. This means that training-time evaluation curves showing the average return are doing so in the DR-limited environments.

    \subsection{Limiting domain randomization in Isaac-Reach-Franka-v0}

        We test three settings of DR, with index one in \autoref{tab:franka-reach-DR-limit-ranges} representing the normal DR ranges for this task from the IsaacLab configuration files.\footnote{Franka Reach configurations \href{https://github.com/isaac-sim/IsaacLab/blob/main/source/isaaclab_tasks/isaaclab_tasks/manager_based/manipulation/reach/reach_env_cfg.py\#L70}{available here.}} We gradually limit the boxes for both the arm's randomly selected target position and its randomly selected starting position. By index 3, the target positions are limited to a small box, the yaw target barely changes the wrist position of the robot, and the reset position barely changes.

        \begin{table}[h]
            \centering
            \caption{Command (target position of the arm, first 4 columns) and reset-randomization (where the arm can start, last column) ranges swept for Isaac-Franka-Reach-v0. Each index corresponds to one of the three experiments.}
            \label{tab:franka-reach-DR-limit-ranges}
            \begin{tabular}{c cccc c}
            \toprule
            Index & $x$ (m) & $y$ (m) & $z$ (m) & yaw (rad) & Joint pos. scale \\
            \midrule
            DR 1 (Default) & $[0.35, 0.65]$ & $[-0.20, 0.20]$ & $[0.15, 0.50]$ & $[-3.14, 3.14]$ & $[0.50, 1.50]$ \\
            DR 2 (Limited) & $[0.45, 0.65]$ & $[-0.10, 0.20]$ & $[0.25, 0.50]$ & $[-1.50, 1.50]$ & $[0.50, 1.00]$ \\
            DR 3 (Severe) & $[0.55, 0.65]$ & $[0.00, 0.20]$  & $[0.35, 0.50]$ & $[-0.50, 0.50]$ & $[0.50, 0.75]$ \\
            \bottomrule
            \end{tabular}
        \end{table}



        \autoref{tab:dr_limits_reach} lists the evaluation performance of DR limited agents on the non-limited task. As the table shows, performance drastically drops as the DR index increases, indicating that domain randomisation improves task generalisation.

        \begin{table}[ht]
        \centering
        \caption{Reach: mean $\pm$ standard deviation return over full episodes (pooled over 5 to 10 seeds and 32 envs per cell), where all models are evaluated on the fully randomised environments, DR 1.}
        \label{tab:dr_limits_reach}
        \begin{tabular}{llccc}
        \toprule
        Agent & $\omega$ & DR 1 (Default) & DR 2 (Limited) & DR 3 (Severe) \\
        \midrule
        \multirow{3}{*}{SAC} & 400 & 0.52 $\pm$ 0.26 & -0.05 $\pm$ 0.53 & -1.18 $\pm$ 0.63 \\
         & 800 & 0.52 $\pm$ 0.25 & -0.18 $\pm$ 0.53 & -1.31 $\pm$ 0.61 \\
         & 1200 & 0.56 $\pm$ 0.23 & 0.04 $\pm$ 0.42 & -1.46 $\pm$ 0.74 \\
        \midrule
        \multirow{3}{*}{\ourMethod} & 400 & 0.52 $\pm$ 0.23 & 0.17 $\pm$ 0.23 & -1.64 $\pm$ 0.71 \\
         & 800 & 0.52 $\pm$ 0.25 & 0.03 $\pm$ 0.36 & -1.49 $\pm$ 0.73 \\
         & 1200 & 0.53 $\pm$ 0.23 & 0.08 $\pm$ 0.24 & -1.57 $\pm$ 0.75 \\
        \midrule
        FastTD3 & N/A & 0.47 $\pm$ 0.26 & -0.26 $\pm$ 0.48 & -1.69 $\pm$ 0.73 \\
        \midrule
        PPO & N/A & 0.62 $\pm$ 0.22 & 0.13 $\pm$ 0.31 & -1.40 $\pm$ 0.61 \\
        \bottomrule
        \end{tabular}
        \end{table}


    \subsection{Limiting domain randomisation in Isaac-Lift-Cube-Franka-v0}

        We again evaluate three DR limitations in Franka Lift, with similar restrictions to Reach, where the first DR index in \autoref{tab:franka-lift-cube-DR-limit-ranges} is the default randomisation from the IsaacLab configuration files.\footnote{Lift Cube configurations \href{https://github.com/isaac-sim/IsaacLab/blob/main/source/isaaclab_tasks/isaaclab_tasks/manager_based/manipulation/lift/lift_env_cfg.py}{available here}} We gradually limit the boxes for both the arm's randomly selected target position and the randomly selected starting position of the cube. By index 3, the target positions are limited to a small box and the object reset position barely changes.

        \begin{table}[h]
            \centering
            \caption{Command (target position of the arm, first 3 columns) and reset-randomization (where the object can start, last two columns) ranges swept for Isaac-Lift-Cube-Franka-v0. Each index corresponds to one of the three experiments.}
            \label{tab:franka-lift-cube-DR-limit-ranges}
            \begin{tabular}{c cccc c}
            \toprule
            Index & target $x$ (m) & target $y$ (m) & target $z$ (m) & object $x$ (m) & object $y$ (m) \\
            \midrule
            DR 1 (Default) & $[0.4, 0.6]$ & $[-0.25, 0.25]$ & $[0.25, 0.50]$ & $[-0.1, 0.1]$ & $[-0.25, 0.25]$ \\
            DR 2 (Limited) & $[0.475, 0.6]$ & $[-0.1, 0.25]$ & $[0.35, 0.50]$ & $[-0.025, 0.1]$ & $[-0.05, 0.25]$ \\
            DR 3 (Severe) & $[0.55, 0.6]$ & $[0.05, 0.25]$ & $[0.45, 0.50]$ & $[0.05, 0.1]$ & $[0.15, 0.25]$ \\
            \bottomrule
            \end{tabular}
        \end{table}

        \begin{table}[ht]
        \centering
        \caption{Lift-Cube: mean $\pm$ standard deviation return over full episodes (pooled over 5 to 10 seeds and 32 envs per cell), where all models are evaluated on the fully randomised environment, DR 1}
        \label{tab:dr_limits_lift_cube}
        \begin{tabular}{llccc}
        \toprule
        Agent & $\omega$ & DR 1 & DR 2 & DR 3 \\
        \midrule
        \multirow{3}{*}{SAC} & 400 & 139.17 $\pm$ 37.56 & 59.23 $\pm$ 73.42 & 62.20 $\pm$ 65.76 \\
         & 800 & 94.20 $\pm$ 76.37 & 90.80 $\pm$ 74.79 & 64.21 $\pm$ 67.76 \\
         & 1200 & 157.67 $\pm$ 2.57 & 127.06 $\pm$ 50.64 & 47.21 $\pm$ 63.69 \\
        \midrule
        \multirow{3}{*}{\ourMethod} & 400 & 115.00 $\pm$ 65.88 & 111.52 $\pm$ 64.29 & 52.96 $\pm$ 63.27 \\
         & 800 & 152.56 $\pm$ 3.44 & 108.16 $\pm$ 63.19 & 73.16 $\pm$ 62.76 \\
         & 1200 & 153.05 $\pm$ 4.88 & 107.34 $\pm$ 66.51 & 39.61 $\pm$ 58.51 \\
         \midrule
        FastTD3 & N/A & 140.47 $\pm$ 6.94 & 125.09 $\pm$ 23.66 & 69.18 $\pm$ 55.33 \\
        \midrule
        PPO & N/A & 136.75 $\pm$ 45.60 & 92.26 $\pm$ 64.24 & 22.76 $\pm$ 31.43 \\
        \bottomrule
        \end{tabular}
        \end{table}

    \subsection{Limiting domain randomisation in Isaac ANYmal-D tasks}
        We evaluate three distinct DR levels, where Level 1 corresponds to the full DR distribution specified in the baseline IsaacLab environment configurations, with added randomisation over friction.\footnote{ANYmal-D configurations \href{https://github.com/isaac-sim/IsaacLab/blob/release/3.0.0-beta2/source/isaaclab_tasks/isaaclab_tasks/manager_based/locomotion/velocity/velocity_env_cfg.py}{available here.}} 
        We restrict physical dynamics and external disturbance ranges by linearly interpolating parameters from full randomization at Level 1 down to nominal, zero-perturbation conditions at Level 3. Specifically, we contract the bounds for surface friction (static and dynamic coefficients), added base mass, and external linear push velocities applied to the robot's base. By Level 3, the friction coefficients are fixed at their nominal values, the added-mass perturbation is fixed to zero, and the periodic push disturbance is completely disabled.

        Independently, we scale the magnitude of additive uniform observation noise applied to the policy's proprioceptive inputs — base linear velocity, base angular velocity, projected gravity, joint position, and joint velocity — halving each noise bound at every successive DR level, so that Level 3 noise is one quarter of its full, Level-1 magnitude.
        
        We further restrict the distribution of initial states sampled at each episode reset using the same linear-interpolation scheme as the dynamics terms, contracting the robot's initial base position, heading, and linear/angular velocity, together with the initial joint-position scale, toward their nominal, non-randomized values. By Level 3, every episode begins from an identical, fully deterministic nominal state, eliminating reset-state diversity entirely so as to induce a pronounced train/eval distribution shift when the resulting policy is evaluated under full randomization.
        
        The exact parameter ranges swept across each level are summarized in \autoref{tab:anymal-d-DR-limit-ranges}.
    
        \begin{table}[ht]
        \centering
        \caption{Anymal-D domain-randomization (DR) limits by level. Observation noise ranges are additive uniform noise on the policy input; reset ranges control the initial state distribution at episode start.}
        \label{tab:anymal-d-DR-limit-ranges}
        \begin{tabular}{lccc}
        \toprule
        Quantity & DR 1 (Default) & DR 2 (Limited) & DR 3 (Severe) \\
        \midrule
        \multicolumn{4}{l}{\textit{Observation noise (additive uniform)}} \\
        Base linear velocity & $\pm$0.10 m/s & $\pm$0.05 m/s & $\pm$0.025 m/s \\
        Base angular velocity & $\pm$0.20 rad/s & $\pm$0.10 rad/s & $\pm$0.05 rad/s \\
        Projected gravity & $\pm$0.05 & $\pm$0.025 & $\pm$0.0125 \\
        Joint position & $\pm$0.01 rad & $\pm$0.005 rad & $\pm$0.0025 rad \\
        Joint velocity & $\pm$1.5 rad/s & $\pm$0.75 rad/s & $\pm$0.375 rad/s \\
        \midrule
        \multicolumn{4}{l}{\textit{Initial state (reset) randomization}} \\
        Base position (x, y) & $\pm$0.50 m & $\pm$0.25 m & 0 m (fixed) \\
        Base heading (yaw) & $\pm\pi$ rad & $\pm\pi/2$ rad & 0 rad (fixed) \\
        Base velocity (all 6 DOF) & $\pm$0.50 (m/s, rad/s) & $\pm$0.25 (m/s, rad/s) & 0 (fixed) \\
        Joint position scale ($\times$ default) & $\times[0.5, 1.5]$ & $\times[0.75, 1.25]$ & $\times1.0$ (fixed) \\
        \midrule
        \multicolumn{4}{l}{\textit{Physics-event randomization}} \\
        Static friction coefficient & $[0.20, 1.25]$ & $[0.50, 1.025]$ & $0.8$ (fixed) \\
        Dynamic friction coefficient & $[0.10, 1.00]$ & $[0.35, 0.80]$ & $0.6$ (fixed) \\
        Added base mass multiplier & $[0.80, 1.25]$ & $[0.90, 1.125]$ & $1.0$ (fixed) \\
        Push-robot velocity kick & $\pm$0.50 m/s & $\pm$0.25 m/s & 0 (fixed) \\
        \bottomrule
        \end{tabular}
        \end{table}

    \begin{table}[ht]
    \centering
    \caption{AnymalD-Flat: mean $\pm$ standard deviation return over full episodes (pooled over 5 to 10 seeds and 32 envs per cell), where all models are evaluated on the fully randomised environments, DR 1.}
    \label{tab:dr_limits_anymald_flat}
    \begin{tabular}{llccc}
    \toprule
    Agent & $\omega$ & DR 1 & DR 2 & DR 3 \\
    \midrule
    \multirow{3}{*}{SAC} & 400 & 24.75 $\pm$ 1.67 & 19.87 $\pm$ 8.58 & 9.48 $\pm$ 13.51 \\
     & 800 & 25.35 $\pm$ 2.09 & 19.54 $\pm$ 10.17 & 4.93 $\pm$ 16.92 \\
     & 1200 & 25.29 $\pm$ 2.82 & 17.48 $\pm$ 12.99 & -0.17 $\pm$ 22.83 \\
    \midrule
    \multirow{3}{*}{\ourMethod} & 400 & 24.60 $\pm$ 2.31 & 20.90 $\pm$ 7.93 & 9.53 $\pm$ 12.07 \\
     & 800 & 25.64 $\pm$ 1.25 & 19.02 $\pm$ 9.96 & 8.98 $\pm$ 12.15 \\
     & 1200 & 25.83 $\pm$ 0.94 & 18.30 $\pm$ 11.06 & 3.07 $\pm$ 19.64 \\
     \midrule
     FastTD3 & N/A & 25.36 $\pm$ 2.56 & 18.37 $\pm$ 13.47 & 1.78 $\pm$ 21.37 \\
     \midrule
     PPO & N/A & 21.73 $\pm$ 6.42 & 10.35 $\pm$ 16.57 & -1.51 $\pm$ 22.86 \\
    \bottomrule
    \end{tabular}
    \end{table}

    \begin{table}[ht]
    \centering
    \caption{AnymalD-Rough: mean $\pm$ standard deviation return over full episodes (pooled over 5 to 10 seeds and 32 envs per cell), where all models are evaluated on the fully randomised environments, DR 1.}
    \label{tab:dr_limits_anymald_rough}
    \begin{tabular}{llccc}
    \toprule
    Agent & $\omega$ & DR 1 & DR 2 & DR 3 \\
    \midrule
    \multirow{3}{*}{SAC} & 400 & 18.04 $\pm$ 6.75 & 20.85 $\pm$ 5.80 & 18.59 $\pm$ 7.39 \\
     & 800 & 21.13 $\pm$ 4.56 & 19.96 $\pm$ 6.26 & 18.48 $\pm$ 7.42 \\
     & 1200 & 19.61 $\pm$ 4.83 & 20.54 $\pm$ 5.55 & 17.76 $\pm$ 8.88 \\
    \midrule
    \multirow{3}{*}{\ourMethod} & 400 & 17.91 $\pm$ 7.21 & 20.18 $\pm$ 6.51 & 18.19 $\pm$ 7.08 \\
     & 800 & 20.57 $\pm$ 4.87 & 20.21 $\pm$ 5.75 & 17.71 $\pm$ 8.32 \\
     & 1200 & 19.96 $\pm$ 3.88 & 20.71 $\pm$ 4.91 & 17.45 $\pm$ 8.39 \\
     \midrule
    FastTD3 & N/A & 10.91 $\pm$ 5.45 & 7.26 $\pm$ 7.17 & 5.70 $\pm$ 8.40 \\
    \midrule
    PPO & N/A & 12.90 $\pm$ 7.14 & 14.52 $\pm$ 7.15 & 13.17 $\pm$ 11.22 \\
    \bottomrule
    \end{tabular}
    \end{table}

\input{robotics_app}

\end{document}

%% file: math_commands.tex
\usepackage{amsmath,amsfonts,bm}

\def\eqref#1{equation~\ref{#1}}

\def\1{\bm{1}}

\DeclareMathAlphabet{\mathsfit}{\encodingdefault}{\sfdefault}{m}{sl}
\SetMathAlphabet{\mathsfit}{bold}{\encodingdefault}{\sfdefault}{bx}{n}



%% file: robotics.tex
\vspace{-6pt}
Our goal is to test whether regularisation makes robotic policies more robust to the sim-to-real gap and to perturbations after deployment. While \autoref{sec:mujoco} demonstrates that \ourMethod learns competitive policies in simulation, the question of real-world transfer robustness remains. 
In this section, we deploy policies trained in simulation on two robots. 
We begin with evaluation on a simulation-to-real matched setup, then perturb individual deployment parameters (sensor noise, object mass, surface friction) to
 assess robustness under domain shift.
\autoref{sec:robotic_nominal} evaluates sim-to-real transfer, and \autoref{sec:robotic_perturb} evaluates robustness to these deployment-time perturbations.
\ourMethod achieves the highest return under every perturbation we tested, supporting our hypothesis that regularised policies would be the most robust to the sim-to-real gap and perturbations.

\subsection{Experimental setup}
\label{sec:robotic_setup}

\begin{figure}[t]
\centering
\includegraphics[width=\textwidth]{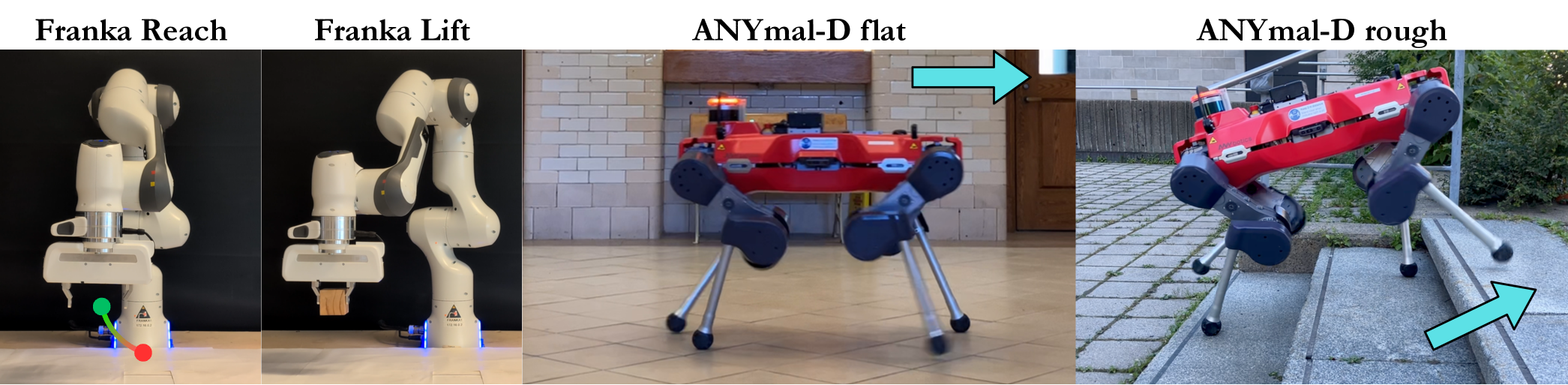}
\caption{The four real-world evaluation settings: Franka Reach and Lift, and ANYmal-D velocity tracking on flat and rough terrain. For the ANYmal-D tasks, blue arrows indicate the robot's direction of commanded velocity.}
\label{fig:robot_setup}
\end{figure}

\textbf{Tasks and platforms}. We evaluate four tasks with two robots: a manipulator and a quadruped. On the Franka Panda arm, the Reach task requires moving the end effector to a target pose, while the Lift task requires grasping a cube and lifting it to a target position. On an ANYmal-D quadruped, we evaluate velocity tracking on flat and rough terrains.
The flat-terrain policy is blind to exteroceptive information, whereas the rough-terrain policy receives perceptual observations. Policies are trained in the corresponding \texttt{Isaac-Reach-Franka-v0}, \texttt{Isaac-Lift-Cube-Franka-v0}, \texttt{Isaac-Velocity-Flat-Anymal-D-v0}, and \texttt{Isaac-Velocity-Rough-Anymal-D-v0} environments using the default DR ranges specified in \aref{sec:isaac_dr}.
\autoref{fig:robot_setup} shows the four deployment settings.

\textbf{Algorithms and configurations}.
We compare PPO, FastTD3, SAC, and \ourMethod.
PPO is implemented with RSL-RL \citep{schwarke2025rslrl}, with sweeps over its entropy coefficient, learning rate, target KL, and rollout length.
FastTD3 is a strong unregularised off-policy baseline \citep{seo2025fasttd3}.
We report both at their best configuration per task.
SAC and \ourMethod use the FlashSAC implementation \citep{kim2026flashsac}. We refer to these simply as SAC and \ourMethod below.
For both regularised methods, we fix $\alpha=1$ and sweep $\omega \in \{400,800,1200\}$.
We do not use SAC temperature autotuning, since doing so would remove the controlled variation in regularisation strength that is central to this experiment.
Each configuration is trained with five random seeds.

\textbf{Reward scales}.
The reward functions have substantially different scales across tasks, with a good episode returning roughly $0.7$ on Reach, $150$ on Lift, and $20$ on either ANYmal task.
Because the regularisation term in the value function is scaled by $1/\omega$ (\autoref{equ:regularised_evaluation_funcs}), a fixed inverse temperature induces different effective regularisation strengths across tasks.
We therefore sweep $\omega$ separately for each task and report all three values.

\textbf{Perturbations}.
To test robustness, we perturb task parameters while keeping each trained policy fixed. The policy, command, and low-level controller are unchanged throughout. 
On Franka, we add Gaussian observation noise ($\sigma=0.005$ on Reach, and $\sigma=0.001$ and $\sigma=0.005$ on Lift). On Lift, we additionally use a cube of twice the training mass, or a cube with a $15$mm center-of-mass (CoM) offset. 
On ANYmal, we lower surface friction by deploying on wet concrete, push the base (flat terrain only), add a $7$kg payload, and add Gaussian noise ($\sigma=0.001$) to the proprioceptive inputs at every step. For safety reasons, a configuration is deployed under a perturbation only if it fails fewer than $80\%$ of simulated episodes under the same perturbation. This rule is fixed in advance and applied to every configuration. As a result, PPO and FastTD3 are excluded from all ANYmal perturbation experiments.
These perturbations are similar to \citet{danesh2025safe}, who evaluate ANYmal-D on surfaces with different friction and with a $3$kg payload.

\textbf{Metrics}.
We report the mean and one standard deviation of the episode return across seeds. \autoref{tab:robot_returns} lists the return of every configuration without perturbation and under each perturbation. \autoref{fig:robot_retention} shows the return under each perturbation as a percentage of the same configuration's return without it. Therefore, a value of $100\%$ indicates that the perturbation incurs no performance cost. Note that values slightly above $100\%$ indicate no loss rather than an improvement. In \aref{sec:robot_appendix}, we additionally report the task error without perturbation (\autoref{tab:robot_nominal}), measured as the distance between the end effector or cube and the target at the end of the episode for Franka, and as the root mean squared error between commanded and measured base velocity for ANYmal.
An ANYmal episode is considered successful if the robot remains upright for the full $20$-second episode.

\begin{table}[t]
\centering
\caption{Return on hardware with and without perturbations.
Entries are the mean and one standard deviation of the return across seeds. Each column is one trained configuration, and reading down a column within a task shows how its return changes under each perturbation. The \textit{none} row is the deployment without perturbation. PPO and FastTD3 are shown at their best configuration per task. \colorbox{highlight}{Highlighted} entries indicate the highest return in each row. Entries marked with ``---'' were not deployed, because the configuration failed the simulation safety check  (\autoref{sec:robotic_setup}). Task errors for the \textit{none} rows are reported in \autoref{tab:robot_nominal}.}
\label{tab:robot_returns}
\small
\setlength{\tabcolsep}{4pt}
\resizebox{\textwidth}{!}{\begin{tabular}{lcccccccc}
\toprule
 & & & \multicolumn{3}{c}{SAC} & \multicolumn{3}{c}{\ourMethod (ours)} \\
\cmidrule(lr){4-6}\cmidrule(lr){7-9}
Perturbation & PPO & FastTD3 & $\omega{=}400$ & $800$ & $1200$ & $\omega{=}400$ & $800$ & $1200$ \\
\midrule
\multicolumn{9}{l}{\textit{Franka Reach}} \\
none & \cellcolor{highlight}0.69 \sd{0.04} & 0.40 \sd{0.09} & 0.48 \sd{0.12} & 0.58 \sd{0.09} & 0.53 \sd{0.09} & 0.51 \sd{0.08} & 0.61 \sd{0.03} & 0.45 \sd{0.05} \\
\quad obs.\ noise ($\sigma{=}0.005$) & 0.19 \sd{0.05} & 0.24 \sd{0.03} & 0.35 \sd{0.03} & 0.40 \sd{0.03} & 0.35 \sd{0.02} & 0.45 \sd{0.02} & \cellcolor{highlight}0.56 \sd{0.02} & 0.33 \sd{0.01} \\
\midrule
\multicolumn{9}{l}{\textit{Franka Lift}} \\
none & 133.1 \sd{54.8} & 123.1 \sd{4.1} & 92.9 \sd{83.8} & 125.8 \sd{69.9} & 94.9 \sd{86.0} & 148.5 \sd{8.6} & \cellcolor{highlight}152.9 \sd{2.5} & 123.9 \sd{69.4} \\
\quad cube mass $2\times$ & 97.5 \sd{11.0} & 102.9 \sd{9.5} & 81.9 \sd{5.2} & 99.1 \sd{4.4} & 87.1 \sd{5.4} & \cellcolor{highlight}152.5 \sd{9.1} & 151.6 \sd{3.1} & 128.3 \sd{8.2} \\
\quad CoM offset ($15$\,mm) & 118.2 \sd{5.6} & 106.8 \sd{3.9} & 93.4 \sd{4.5} & 115.0 \sd{7.8} & 96.2 \sd{5.9} & \cellcolor{highlight}154.4 \sd{9.2} & 153.0 \sd{2.8} & 127.6 \sd{10.3} \\
\quad obs.\ noise ($\sigma{=}0.001$) & 38.0 \sd{8.4} & 96.5 \sd{3.0} & 81.1 \sd{4.8} & 103.4 \sd{11.6} & 82.1 \sd{5.5} & 147.0 \sd{3.1} & \cellcolor{highlight}150.2 \sd{2.3} & 122.4 \sd{2.5} \\
\quad obs.\ noise ($\sigma{=}0.005$) & 19.4 \sd{9.2} & 72.4 \sd{5.7} & 60.7 \sd{7.2} & 68.5 \sd{7.8} & 62.5 \sd{6.6} & \cellcolor{highlight}120.9 \sd{12.2} & \cellcolor{highlight}120.9 \sd{4.0} & 107.7 \sd{4.6} \\
\midrule
\multicolumn{9}{l}{\textit{ANYmal flat}} \\
none & 18.9 \sd{5.6} & 23.6 \sd{0.7} & 25.3 \sd{0.3} & 25.2 \sd{0.5} & 24.8 \sd{0.1} & 25.0 \sd{0.4} & \cellcolor{highlight}25.9 \sd{0.4} & 25.5 \sd{0.3} \\
\quad low friction & --- & --- & 22.3 \sd{1.3} & 22.5 \sd{1.1} & 21.5 \sd{1.5} & 24.5 \sd{0.2} & \cellcolor{highlight}24.7 \sd{0.4} & 24.6 \sd{0.3} \\
\quad push & --- & --- & 23.0 \sd{1.2} & 22.8 \sd{1.0} & 21.8 \sd{1.4} & \cellcolor{highlight}25.3 \sd{0.2} & \cellcolor{highlight}25.3 \sd{0.3} & 24.9 \sd{0.3} \\
\quad payload & --- & --- & 24.1 \sd{0.7} & 22.6 \sd{1.1} & 22.8 \sd{0.9} & 25.1 \sd{0.2} & \cellcolor{highlight}25.2 \sd{0.3} & 25.0 \sd{0.2} \\
\quad proprioceptive noise & --- & --- & 20.7 \sd{2.2} & 20.5 \sd{2.0} & 20.5 \sd{1.9} & \cellcolor{highlight}24.7 \sd{0.2} & \cellcolor{highlight}24.7 \sd{0.4} & \cellcolor{highlight}24.7 \sd{0.3} \\
\midrule
\multicolumn{9}{l}{\textit{ANYmal rough}} \\
none & 8.6 \sd{3.7} & 13.7 \sd{4.1} & 19.1 \sd{1.6} & 19.3 \sd{4.9} & 18.8 \sd{3.4} & 21.0 \sd{1.2} & 20.8 \sd{1.3} & \cellcolor{highlight}21.9 \sd{1.6} \\
\quad low friction & --- & --- & 16.4 \sd{1.5} & 17.0 \sd{0.8} & 17.4 \sd{1.0} & 18.5 \sd{0.6} & 21.2 \sd{0.6} & \cellcolor{highlight}21.9 \sd{1.1} \\
\quad payload & --- & --- & 17.2 \sd{0.6} & 16.8 \sd{0.9} & 16.8 \sd{1.9} & 21.2 \sd{0.8} & 20.5 \sd{0.5} & \cellcolor{highlight}21.5 \sd{1.1} \\
\quad proprioceptive noise & --- & --- & 15.3 \sd{1.6} & 15.3 \sd{1.9} & 13.5 \sd{1.0} & 18.8 \sd{1.4} & \cellcolor{highlight}19.3 \sd{0.7} & 18.9 \sd{1.9} \\
\bottomrule
\end{tabular}}
\end{table}

\subsection{Sim-to-real transfer}
\label{sec:robotic_nominal}

\textbf{\ourMethod transfers best on Lift and both locomotion tasks.}
The \textit{none} rows of \autoref{tab:robot_returns} report the return of each policy deployed without perturbation. In locomotion, every regularised configuration achieves a higher return that PPO and FastTD3 on both terrains, with the largest gap on rough terrain. 
On Lift, the two best configurations are both \ourMethod.
The best of these, at $\omega=800$, is also far more consistent across seeds than PPO and SAC, whose large standard deviations indicate that their transfer varies considerably from seed to seed.

\textbf{Regularised policies remain deployable under a larger sim-to-real gap.}
We further widen the sim-to-real gap by limiting domain randomisation in training (\aref{sec:isaac_dr}) and deploying the resulting policies on hardware (\aref{sec:robot_dr_appendix}).
Limiting domain randomisation lowers the return of every method, indicating that DR remains important for regularised policies as well.
On ANYmal, however, PPO and FastTD3 no longer pass the simulation safety check once the randomisation is limited, and are thus not deployable.
On Lift, \ourMethod achieves the highest return at every DR level.
Regularisation and DR thus appear complementary, since limiting the randomisation hurts every method, while regularised policies remain deployable where unregularised ones are not.


\subsection{Robustness to deployment-time perturbations}
\label{sec:robotic_perturb}

\begin{figure}[t]
\centering
\includegraphics[width=\textwidth]{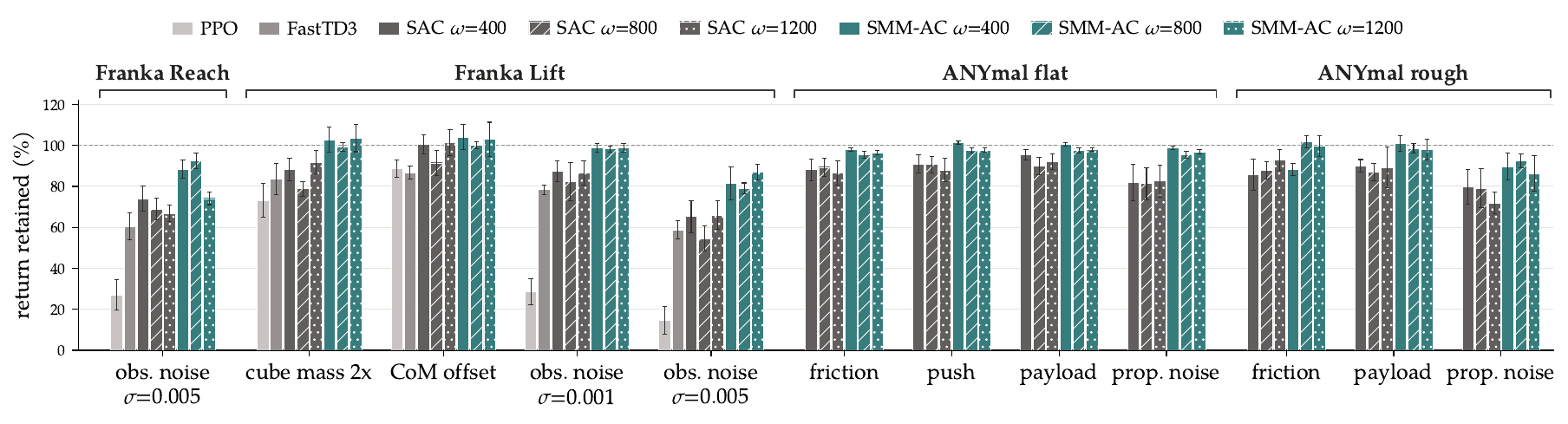}
\caption{Return kept under real-world deployment-time perturbations. Return under each perturbation as a percentage of the same configuration's return without it, so $100\%$ means no loss. Bars are seed means, and error bars show one standard deviation. Franka is perturbed with Gaussian observation noise and, on Lift, with a cube of 2$\times$ the training mass and with a $15$mm CoM offset. ANYmal is perturbed with lower surface friction, pushes, a payload, and proprioceptive noise. PPO and FastTD3 do not appear on ANYmal because they fail the simulation safety check, which is applied identically to every configuration.}
\label{fig:robot_retention}
\end{figure}

We now perturb task parameters and report the resulting return as a percentage of its unperturbed return in \autoref{fig:robot_retention}.
\autoref{tab:robot_returns} reports the corresponding absolute returns.

\textbf{\ourMethod keeps more of its return than the baselines.}
At the same $\omega$, \ourMethod keeps more of its return than SAC under every perturbation, in most cases by more than two standard deviations combined.
It retains more than PPO and FastTD3 wherever those were deployed, in every case by more than the combined standard deviations.
Its results are also more consistent across seeds, with about half the standard deviation of SAC on average. Under most perturbations, \ourMethod loses little or none of its return, while the baselines retain considerably less. However, the empirical drop in performance depends on the type of perturbation.

\textbf{Observation noise causes the largest losses.}
Every method has its largest loss under observation noise, and this is also where the methods differ the most. On Franka, PPO loses most of its return under noise, and FastTD3 and SAC lose a substantial part of theirs. Nevertheless, \ourMethod keeps most of its return.
On ANYmal, proprioceptive noise separates SAC and \ourMethod more than any change to the physical system. Since the dynamics are untouched, these losses come entirely from how each policy responds to corrupted inputs.

\textbf{Changes to the physical system cost less.}
A heavier or unbalanced cube costs \ourMethod nothing, while baselines lose part of their returns.
On ANYmal, \ourMethod loses little under lower friction, pushes, and payload on either terrain, while SAC loses noticeably more.
The one exception is lower friction on rough terrain at $\omega=400$ where \ourMethod loses about as much as SAC. At other values of $\omega$, it loses nothing.

\textbf{\ourMethod maintains the highest absolute return.}
Note that a policy with a low return can still keep a high percentage of it. Therefore, \autoref{tab:robot_returns} reports absolute returns as well. 
Under every perturbation, the highest return in achieved by \ourMethod.
On Lift and locomotion tasks, every configuration of \ourMethod outperforms every deployed baseline under every perturbation.
On Reach, under observation noise, \ourMethod at $\omega=800$ achieves the highest return and PPO the lowest.

\textbf{Discussion.}
These experiments separate two effects that are easy to conflate, namely transferring well to hardware, and keeping that performance as conditions move away from training.
Regularisation helps most clearly with the second.
Across all perturbations, \ourMethod keeps the most return and achieves the highest absolute return.
Its advantage is largest under observation noise, which every real robot faces to some degree in real-world deployment. These results support our hypothesis that regularisation improves sim-to-real performance and robustness to perturbations. They also empirically validate theoretical robustness claims, providing evidence that SMM, with its larger robust sets, is indeed more robust.

%% file: robotics_app.tex
\section{Additional hardware results}
\label{sec:robot_appendix}

\autoref{tab:robot_nominal} extends the \textit{none} rows of \autoref{tab:robot_returns} with the task error of each configuration.
The errors follow the returns, and the configuration with the highest return on a task also has the lowest error on Reach, Lift, and both ANYmal tasks.

\begin{table}[t]
\centering
\caption{Hardware deployment without perturbation: task error and return, mean and one standard deviation over seeds.
Each column is one trained configuration. On Franka, the error is the distance between the end effector (Reach) or the cube (Lift) and the target at the end of the episode. On ANYmal, it is the RMSE between commanded and measured base velocity.
PPO and FastTD3 are shown at their best configuration per task. \colorbox{highlight}{Highlighted} entries indicate the best mean within each task metric.}
\label{tab:robot_nominal}
\small
\setlength{\tabcolsep}{4pt}
\resizebox{\textwidth}{!}{\begin{tabular}{lcccccccc}
\toprule
 & & & \multicolumn{3}{c}{SAC} & \multicolumn{3}{c}{\ourMethod (ours)} \\
\cmidrule(lr){4-6}\cmidrule(lr){7-9}
Task & PPO & FastTD3 & $\omega{=}400$ & $800$ & $1200$ & $\omega{=}400$ & $800$ & $1200$ \\
\midrule
\multicolumn{9}{l}{\textit{Franka Reach}} \\
Error (cm)
& \cellcolor{highlight}0.64 \sd{0.11}
& 0.93 \sd{0.14}
& 0.93 \sd{0.16}
& 0.79 \sd{0.12}
& 0.91 \sd{0.15}
& 1.00 \sd{0.15}
& 0.76 \sd{0.10}
& 1.07 \sd{0.17} \\
Return
& \cellcolor{highlight}0.69 \sd{0.04}
& 0.40 \sd{0.09}
& 0.48 \sd{0.12}
& 0.58 \sd{0.09}
& 0.53 \sd{0.09}
& 0.51 \sd{0.08}
& 0.61 \sd{0.03}
& 0.45 \sd{0.05} \\
\midrule
\multicolumn{9}{l}{\textit{Franka Lift}} \\
Error (cm)
& 5.79 \sd{3.6}
& 4.26 \sd{0.6}
& 16.53 \sd{9.4}
& 8.54 \sd{7.1}
& 17.32 \sd{9.7}
& 1.53 \sd{0.6}
& \cellcolor{highlight}0.82 \sd{0.2}
& 8.57 \sd{6.6} \\
Return
& 133.10 \sd{54.75}
& 123.14 \sd{4.05}
& 92.90 \sd{83.80}
& 125.78 \sd{69.93}
& 94.93 \sd{86.04}
& 148.48 \sd{8.61}
& \cellcolor{highlight}152.87 \sd{2.52}
& 123.94 \sd{69.39} \\
\midrule
\multicolumn{9}{l}{\textit{ANYmal flat}} \\
RMSE (m/s)
& 0.2289 \sd{0.047}
& 0.1955 \sd{0.018}
& 0.1439 \sd{0.012}
& 0.1369 \sd{0.010}
& 0.1355 \sd{0.009}
& 0.1509 \sd{0.011}
& \cellcolor{highlight}0.1268 \sd{0.008}
& 0.1523 \sd{0.012} \\
Return
& 18.94 \sd{5.55}
& 23.60 \sd{0.70}
& 25.28 \sd{0.28}
& 25.18 \sd{0.45}
& 24.77 \sd{0.09}
& 24.98 \sd{0.42}
& \cellcolor{highlight}25.89 \sd{0.40}
& 25.51 \sd{0.30} \\
\midrule
\multicolumn{9}{l}{\textit{ANYmal rough}} \\
RMSE (m/s)
& 0.5775 \sd{0.081}
& 0.5295 \sd{0.073}
& 0.3689 \sd{0.044}
& 0.3471 \sd{0.057}
& 0.3386 \sd{0.049}
& 0.4112 \sd{0.045}
& 0.3252 \sd{0.038}
& \cellcolor{highlight}0.2938 \sd{0.034} \\
Return
& 8.60 \sd{3.66}
& 13.72 \sd{4.13}
& 19.11 \sd{1.62}
& 19.33 \sd{4.90}
& 18.78 \sd{3.40}
& 20.96 \sd{1.18}
& 20.82 \sd{1.32}
& \cellcolor{highlight}21.93 \sd{1.60} \\
\bottomrule
\end{tabular}}
\end{table}

\subsection{Hardware deployment with limited domain randomisation}
\label{sec:robot_dr_appendix}

In \aref{sec:isaac_dr}, we train policies with progressively narrower DR (DR 2 and DR 3) and evaluate them in simulation under the default ranges (DR 1).
Here we deploy the same checkpoints on hardware without any perturbation, as in \autoref{sec:robotic_nominal}. This can be viewed as another way of increasing the sim-to-real gap, complementing the experiments in \autoref{sec:robotic_perturb}.
Before deployment, each checkpoint goes through the simulation safety check described in \autoref{sec:robotic_setup}.
This check rules out PPO at DR 2 and PPO and FastTD3 at DR 3 on both ANYmal tasks, as well as every configuration at DR 3 on flat terrain.
\autoref{tab:robot_dr_limited} reports the returns, and its DR 1 column repeats the \textit{none} rows of \autoref{tab:robot_returns}.

\begin{table}[t]
\centering
\caption{Hardware return of policies trained with limited DR, deployed without perturbation.
Each column is one training configuration, and reading down a column within a task shows how its hardware return changes with the training-time DR level of \aref{sec:isaac_dr}. The \textit{DR 1} row corresponds to the \textit{none} rows of \autoref{tab:robot_returns}. Entries are the mean and one standard deviation of the return across seeds. \colorbox{highlight}{Highlighted} entries indicate the highest return in each column and task. Entries marked with ``---'' were not deployed, because the configuration failed the simulation safety check (\autoref{sec:robotic_setup}).}
\label{tab:robot_dr_limited}
\small
\setlength{\tabcolsep}{4pt}
\resizebox{\textwidth}{!}{\begin{tabular}{llcccccccc}
\toprule
 & & & \multicolumn{3}{c}{SAC} & \multicolumn{3}{c}{\ourMethod (ours)} \\
\cmidrule(lr){4-6}\cmidrule(lr){7-9}
DR level & PPO & FastTD3 & $\omega{=}400$ & $800$ & $1200$ & $\omega{=}400$ & $800$ & $1200$ \\
\midrule
\multicolumn{9}{l}{\textit{Franka Reach}} \\
DR 1 (Default) & \cellcolor{highlight}0.69 \sd{0.04} & 0.40 \sd{0.09} & 0.48 \sd{0.12} & 0.58 \sd{0.09} & 0.53 \sd{0.09} & 0.51 \sd{0.08} & 0.61 \sd{0.03} & 0.45 \sd{0.05} \\
DR 2 (Limited) & \cellcolor{highlight}0.24 \sd{0.08} & -0.16 \sd{0.11} & -0.01 \sd{0.14} & -0.11 \sd{0.13} & 0.08 \sd{0.11} & 0.21 \sd{0.07} & 0.12 \sd{0.09} & 0.16 \sd{0.06} \\
DR 3 (Severe) & -1.18 \sd{0.13} & -1.42 \sd{0.18} & \cellcolor{highlight}-1.03 \sd{0.16} & -1.15 \sd{0.15} & -1.27 \sd{0.19} & -1.38 \sd{0.14} & -1.29 \sd{0.16} & -1.34 \sd{0.17} \\
\midrule
\multicolumn{9}{l}{\textit{Franka Lift}} \\
DR 1 (Default) & 133.1 \sd{54.8} & 123.1 \sd{4.1} & 92.9 \sd{83.8} & 125.8 \sd{69.9} & 94.9 \sd{86.0} & 148.5 \sd{8.6} & \cellcolor{highlight}152.9 \sd{2.5} & 123.9 \sd{69.4} \\
DR 2 (Limited) & 98.4 \sd{61.4} & 117.8 \sd{19.3} & 47.6 \sd{67.6} & 112.4 \sd{62.2} & 106.8 \sd{53.3} & \cellcolor{highlight}136.3 \sd{24.6} & 118.5 \sd{49.8} & 93.8 \sd{71.2} \\
DR 3 (Severe) & 16.8 \sd{31.5} & 58.4 \sd{48.6} & 64.2 \sd{63.8} & 71.6 \sd{66.0} & 37.4 \sd{56.9} & 61.9 \sd{59.4} & \cellcolor{highlight}81.2 \sd{58.1} & 31.7 \sd{49.4} \\
\midrule
\multicolumn{9}{l}{\textit{ANYmal flat}} \\
DR 1 (Default) & 18.9 \sd{5.6} & 23.6 \sd{0.7} & 25.3 \sd{0.3} & 25.2 \sd{0.5} & 24.8 \sd{0.1} & 25.0 \sd{0.4} & \cellcolor{highlight}25.9 \sd{0.4} & 25.5 \sd{0.3} \\
DR 2 (Limited) & --- & 16.4 \sd{1.1} & 20.8 \sd{2.0} & 11.4 \sd{4.9} & 17.6 \sd{3.4} & \cellcolor{highlight}22.2 \sd{1.2} & 20.5 \sd{1.8} & 19.2 \sd{2.1} \\
DR 3 (Severe) & --- & --- & --- & --- & --- & --- & --- & --- \\
\midrule
\multicolumn{9}{l}{\textit{ANYmal rough}} \\
DR 1 (Default) & 8.6 \sd{3.7} & 13.7 \sd{4.1} & 19.1 \sd{1.6} & 19.3 \sd{4.9} & 18.8 \sd{3.4} & 21.0 \sd{1.2} & 20.8 \sd{1.3} & \cellcolor{highlight}21.9 \sd{1.6} \\
DR 2 (Limited) & --- & --- & 21.4 \sd{1.9} & 18.6 \sd{3.8} & 19.8 \sd{2.5} & 19.7 \sd{2.1} & 21.2 \sd{1.5} & \cellcolor{highlight}22.0 \sd{1.5} \\
DR 3 (Severe) & --- & --- & 17.7 \sd{2.9} & 16.9 \sd{4.1} & 15.8 \sd{4.6} & \cellcolor{highlight}18.3 \sd{2.6} & 17.4 \sd{3.3} & 18.1 \sd{3.1} \\
\bottomrule
\end{tabular}}
\end{table}

\textbf{Limiting domain randomisation lowers the return on hardware.}
Every configuration deployed at DR 3 achieves a lower return than at DR 1.
On Reach, every DR 3 configuration obtains a negative return, and on flat terrain, no DR 3 configuration passes the safety check at all.
The effect of DR 2 depends on the task.
On Reach and flat terrain, every deployed configuration already loses return at DR 2, and on Lift all but one do.
On rough terrain, the return at DR 2 stays close to DR 1, and several configurations score slightly higher, in line with the simulation results in \autoref{tab:dr_limits_anymald_rough}.

\textbf{\ourMethod has the highest return on Lift and in locomotion.}
On Lift and both ANYmal tasks, the highest return at every deployed DR level comes from a \ourMethod configuration.
Note, however, that the seed-to-seed variation under limited DR is large, and most differences between SAC and \ourMethod at the same $\omega$ are smaller than their combined standard deviations.

\textbf{Discussion.}
Limiting DR costs every method return on hardware, so regularisation does not replace it.
Under limited randomisation, however, the regularised policies still pass the safety check on ANYmal, which the unregularised baselines struggle more.

%% file: ref.bib
@article{TGM_2025_jiralerspong,
  title={Robust Reinforcement Learning for Discrete Compositional Generation via General Soft Operators},
  author={Jiralerspong, Marco and Derman, Esther and Vucetic, Danilo and Malkin, Nikolay and Sun, Bilun and Zhang, Tianyu and Bacon, Pierre-Luc and Gidel, Gauthier},
  journal={arXiv preprint arXiv:2506.17007},
  year={2025}
}

@inproceedings{geist2019theory_regularizers,
  title={A theory of regularized markov decision processes},
  author={Geist, Matthieu and Scherrer, Bruno and Pietquin, Olivier},
  booktitle={International conference on machine learning},
  pages={2160--2169},
  year={2019},
  organization={PMLR}
}

@article{danesh2025safe,
  title={Safe domain randomization via uncertainty-aware out-of-distribution detection and policy adaptation},
  author={Danesh, Mohamad H and Wabartha, Maxime and Wu, Stanley and Pineau, Joelle and Lin, Hsiu-Chin},
  journal={arXiv preprint arXiv:2507.06111},
  year={2025}
}

@inproceedings{asadi2017alternative_mellowmax,
    title={An alternative softmax operator for reinforcement learning},
    author={Asadi, Kavosh and Littman, Michael L},
    booktitle={International Conference on Machine Learning},
    pages={243--252},
    year={2017},
    organization={PMLR}
}

@inproceedings{gan2021stabilizing_soft_mellowmax,
    title={Stabilizing q learning via soft mellowmax operator},
    author={Gan, Yaozhong and Zhang, Zhe and Tan, Xiaoyang},
    booktitle={Proceedings of the AAAI Conference on Artificial Intelligence},
    volume={35},
    pages={7501--7509},
    year={2021}
}

@inproceedings{haarnoja2018soft,
  title={Soft actor-critic: Off-policy maximum entropy deep reinforcement learning with a stochastic actor},
  author={Haarnoja, Tuomas and Zhou, Aurick and Abbeel, Pieter and Levine, Sergey},
  booktitle={International conference on machine learning},
  pages={1861--1870},
  year={2018},
  organization={Pmlr}
}

@article{
   genesis2026genesisworld,
   author = {Genesis, AI Team},
   title = {The Role of Simulation in Scalable Robotics, Genesis World 1.0, and the Path Forward},
   journal = {Genesis AI Blog},
   month = {May},
   year = {2026},
}

@article{haarnoja2018SAC_2,
  title={Soft actor-critic algorithms and applications},
  author={Haarnoja, Tuomas and Zhou, Aurick and Hartikainen, Kristian and Tucker, George and Ha, Sehoon and Tan, Jie and Kumar, Vikash and Zhu, Henry and Gupta, Abhishek and Abbeel, Pieter and others},
  journal={arXiv preprint arXiv:1812.05905},
  year={2018}
}

@inproceedings{todorov2012mujoco,
  title={MuJoCo: A physics engine for model-based control},
  author={Todorov, Emanuel and Erez, Tom and Tassa, Yuval},
  booktitle={2012 IEEE/RSJ International Conference on Intelligent Robots and Systems},
  pages={5026--5033},
  year={2012},
  organization={IEEE},
  doi={10.1109/IROS.2012.6386109}
}

@inproceedings{fujimoto2018addressing_TD3,
  title={Addressing function approximation error in actor-critic methods},
  author={Fujimoto, Scott and Hoof, Herke and Meger, David},
  booktitle={International conference on machine learning},
  pages={1587--1596},
  year={2018},
  organization={PMLR}
}

@article{schulman2017proximal_PPO,
  title={Proximal policy optimization algorithms},
  author={Schulman, John and Wolski, Filip and Dhariwal, Prafulla and Radford, Alec and Klimov, Oleg},
  journal={arXiv preprint arXiv:1707.06347},
  year={2017}
}

@article{Mittal_Isaac_Lab_2025,
author = {Mittal, Mayank and Roth, Pascal and Tigue, James and Richard, Antoine and Zhang, Octi and Du, Peter and Serrano-Muñoz, Antonio and Yao, Xinjie and Zurbrügg, René and Rudin, Nikita and Wawrzyniak, Lukasz and Rakhsha, Milad and Denzler, Alain and Heiden, Eric and Borovicka, Ales and Ahmed, Ossama and Akinola, Iretiayo and Anwar, Abrar and Carlson, Mark T. and Feng, Ji Yuan and Garg, Animesh and Gasoto, Renato and Gulich, Lionel and Guo, Yijie and Gussert, M. and Hansen, Alex and Kulkarni, Mihir and Li, Chenran and Liu, Wei and Makoviychuk, Viktor and Malczyk, Grzegorz and Mazhar, Hammad and Moghani, Masoud and Murali, Adithyavairavan and Noseworthy, Michael and Poddubny, Alexander and Ratliff, Nathan and Rehberg, Welf and Schwarke, Clemens and Singh, Ritvik and Smith, James Latham and Tang, Bingjie and Thaker, Ruchik and Trepte, Matthew and Van Wyk, Karl and Yu, Fangzhou and Millane, Alex and Ramasamy, Vikram and Steiner, Remo and Subramanian, Sangeeta and Volk, Clemens and Chen, CY and Jawale, Neel and Kuruttukulam, Ashwin Varghese and Lin, Michael A. and Mandlekar, Ajay and Patzwaldt, Karsten and Welsh, John and Lafleche, Jean-Francois and Moënne-Loccoz, Nicolas and Park, Soowan and Stepinski, Rob and Van Gelder, Dirk and Amevor, Chris and Carius, Jan and Chang, Jumyung and He Chen, Anka and Ciechomski, Pablo de Heras and Daviet, Gilles and Mohajerani, Mohammad and von Muralt, Julia and Reutskyy, Viktor and Sauter, Michael and Schirm, Simon and Shi, Eric L. and Terdiman, Pierre and Vilella, Kenny and Widmer, Tobias and Yeoman, Gordon and Chen, Tiffany and Grizan, Sergey and Li, Cathy and Li, Lotus and Smith, Connor and Wiltz, Rafael and Alexis, Kostas and Chang, Yan and Fan, Linxi "Jim" and Farshidian, Farbod and Handa, Ankur and Huang, Spencer and Hutter, Marco and Narang, Yashraj and Pouya, Soha and Sheng, Shiwei and Zhu, Yuke and Macklin, Miles and Moravanszky, Adam and Reist, Philipp and Guo, Yunrong and Hoeller, David and State, Gavriel},
doi = {10.48550/arXiv.2511.04831},
journal = {arXiv preprint arXiv:2511.04831},
title = {{Isaac Lab - A GPU-Accelerated Simulation Framework for Multi-Modal Robot Learning}},
url = {https://arxiv.org/abs/2511.04831},
year = {2025}
}

@article{schwarke2025rslrl,
  title={RSL-RL: A Learning Library for Robotics Research},
  author={Schwarke, Clemens and Mittal, Mayank and Rudin, Nikita and Hoeller, David and Hutter, Marco},
  journal={arXiv preprint arXiv:2509.10771},
  year={2025}
}

@article{rslrlsac2026bridging,
  title={Bridging the Gap: Enabling Soft Actor Critic for High Performance Legged Locomotion},
  author={Sabatini, Gianluca and Li, Chenhao and Hutter, Marco},
  journal={arXiv preprint arXiv:2605.24975},
  year={2026}
}

@article{neu2017unified,
  title={A unified view of entropy-regularized markov decision processes},
  author={Neu, Gergely and Jonsson, Anders and G{\'o}mez, Vicen{\c{c}}},
  journal={arXiv preprint arXiv:1705.07798},
  year={2017}
}

@article{kim2026flashsac,
  title={FlashSAC: Fast and Stable Off-Policy Reinforcement Learning for High-Dimensional Robot Control},
  author={Kim, Donghu and Lee, Youngdo and Park, Minho and Kim, Kinam and Nahendra, I Made Aswin and Seno, Takuma and Min, Sehee and Palenicek, Daniel and Vogt, Florian and Kragic, Danica and Peters, Jan and Choo, Jaegul and Lee, Hojoon},
  journal={arXiv preprint arXiv:2604.04539},
  year={2026}
}

@inproceedings{kingma2013auto,
  title={Auto-Encoding Variational Bayes},
  author={Kingma, Diederik P and Welling, Max},
  booktitle={International Conference on Learning Representations, {ICLR}},
  year={2014},
  url={https://arxiv.org/abs/1312.6114}
}

@inproceedings{li2023parallel_PQL,
  title={Parallel $Q$-Learning: Scaling Off-policy Reinforcement Learning under Massively Parallel Simulation},
  author={Li, Zechu and Chen, Tao and Hong, Zhang-Wei and Ajay, Anurag and Agrawal, Pulkit},
  booktitle={International Conference on Machine Learning},
  year={2023},
  organization={PMLR}
}

@inproceedings{derman_twice_regularized,
    author = {Derman, Esther and Geist, Matthieu and Mannor, Shie},
    booktitle = {Advances in Neural Information Processing Systems},
    editor = {M. Ranzato and A. Beygelzimer and Y. Dauphin and P.S. Liang and J. Wortman Vaughan},
    pages = {22274--22287},
    publisher = {Curran Associates, Inc.},
    title = {Twice regularized MDPs and the equivalence between robustness and regularization},
    url = {https://proceedings.neurips.cc/paper_files/paper/2021/file/bb1443cc31d7396bf73e7858cea114e1-Paper.pdf},
    volume = {34},
    year = {2021}
}

@inproceedings{eysenbach2022maximum,
  author    = {Benjamin Eysenbach and
               Sergey Levine},
  title     = {Maximum Entropy {RL} (Provably) Solves Some Robust {RL} Problems},
  booktitle = {Tenth International Conference on Learning Representations, {ICLR}
               2022, Virtual Event, April 25-29, 2022},
  publisher = {OpenReview.net},
  year      = {2022},
  url       = {https://openreview.net/forum?id=PtSAD3caaA2}
}

@inproceedings{zhang2025when,
  author    = {Ruipeng Zhang and
               Ya{-}Chien Chang and
               Sicun Gao},
  title     = {When Maximum Entropy Misleads Policy Optimization},
  booktitle = {Proceedings of the 42nd International Conference on Machine Learning,
               {ICML} 2025, Vancouver, Canada},
  series    = {Proceedings of Machine Learning Research},
  volume    = {267},
  publisher = {{PMLR}},
  year      = {2025}
}

@InProceedings{pmlr-v130-husain21a_reg_pol_rew_rob,
  title = 	 { Regularized Policies are Reward Robust },
  author =       {Husain, Hisham and Ciosek, Kamil and Tomioka, Ryota},
  booktitle = 	 {Proceedings of The 24th International Conference on Artificial Intelligence and Statistics},
  pages = 	 {64--72},
  year = 	 {2021},
  editor = 	 {Banerjee, Arindam and Fukumizu, Kenji},
  volume = 	 {130},
  series = 	 {Proceedings of Machine Learning Research},
  month = 	 {13--15 Apr},
  publisher =    {PMLR},
  url = 	 {https://proceedings.mlr.press/v130/husain21a.html}
}

@article{seo2025fasttd3,
  title={FastTD3: Simple, Fast, and Capable Reinforcement Learning for Humanoid Control},
  author={Seo, Younggyo and Sferrazza, Carmelo and Geng, Haoran and Nauman, Michal and Yin, Zhao-Heng and Abbeel, Pieter},
  journal={arXiv preprint arXiv:2505.22642},
  year={2025}
}

@article{huang2022cleanrl,
      author  = {Shengyi Huang and Rousslan Fernand Julien Dossa and Chang Ye and Jeff Braga and Dipam Chakraborty and Kinal Mehta and João G.M. Araújo},
      title   = {CleanRL: High-quality Single-file Implementations of Deep Reinforcement Learning Algorithms},
      journal = {Journal of Machine Learning Research},
      year    = {2022},
      volume  = {23},
      number  = {274},
      pages   = {1--18},
      url     = {http://jmlr.org/papers/v23/21-1342.html}
}

@article{seo2025learning_fastsac,
  title={Learning Sim-to-Real Humanoid Locomotion in 15 Minutes},
  author={Seo, Younggyo and Sferrazza, Carmelo and Chen, Juyue and Shi, Guanya and Duan, Rocky and Abbeel, Pieter},
  journal={arXiv preprint arXiv:2512.01996},
  year={2025}
}

@inproceedings{gallici2025simplifying_pqn,
  title={Simplifying deep temporal difference learning},
  author={Gallici, Matteo and Fellows, Mattie and Ellis, Benjamin and Pou, Bartomeu and Masmitja, Ivan and Foerster, Jakob and Martin, Mario},
  booktitle={International Conference on Learning Representations},
  volume={2025},
  pages={78148--78190},
  year={2025}
}

@article{dam2024unified_tree_search,
  title={A unified perspective on value backup and exploration in monte-carlo tree search},
  author={Dam, Tuan and D'Eramo, Carlo and Peters, Jan and Pajarinen, Joni},
  journal={Journal of Artificial Intelligence Research},
  volume={81},
  pages={511--577},
  year={2024}
}

@article{brekelmans2022_divergence_robustness,
  title={Your policy regularizer is secretly an adversary},
  author={Brekelmans, Rob and Genewein, Tim and Grau-Moya, Jordi and Del{\'e}tang, Gr{\'e}goire and Kunesch, Markus and Legg, Shane and Ortega, Pedro},
  journal={arXiv preprint arXiv:2203.12592},
  year={2022}
}

@article{gong_KL_actor_critic_prev_policy,
    author = {Gong, Mingrong and Yi, Zhengkun and Chen, Yidong and Li, Huiyun and Cui, Yunduan},
    title = {Reducing the value function over-estimation by Kullback-Leibler divergence regularized distributional actor-critic},
    journal = {Applied Intelligence},
    year    = {2025},
    volume  = {55},
}

@inproceedings{lee2020_tsallis_actor_critic_robotics,
  title={Generalized Tsallis Entropy Reinforcement Learning and Its Application to Soft Mobile Robots.},
  author={Lee, Kyungjae and Kim, Sungyub and Lim, Sungbin and Choi, Sungjoon and Hong, Mineui and Kim, Jae In and Park, Yong-Lae and Oh, Songhwai},
  booktitle={Robotics: science and systems},
  volume={16},
  pages={1--10},
  year={2020}
}

@inproceedings{ahmed_le_roux_2019_entropy_impact_pol_opt,
  title={Understanding the impact of entropy on policy optimization},
  author={Ahmed, Zafarali and Le Roux, Nicolas and Norouzi, Mohammad and Schuurmans, Dale},
  booktitle={International conference on machine learning},
  pages={151--160},
  year={2019},
  organization={PMLR}
}

@article{mock_2023_compare_PPO_TD3_SAC_sim_legged_robo,
    author = {Mock, James and Muknahallipatna, Suresh},
    year = {2023},
    month = {01},
    pages = {36-56},
    title = {A Comparison of PPO, TD3 and SAC Reinforcement Algorithms for Quadruped Walking Gait Generation},
    volume = {15},
    journal = {Journal of Intelligent Learning Systems and Applications},
    doi = {10.4236/jilsa.2023.151003}
}

@inproceedings{mysore2021regularizing,
  title={Regularizing action policies for smooth control with reinforcement learning},
  author={Mysore, Siddharth and Mabsout, Bassel and Mancuso, Renato and Saenko, Kate},
  booktitle={2021 IEEE International Conference on Robotics and Automation (ICRA)},
  pages={1810--1816},
  year={2021},
  organization={IEEE}
}

@article{van2018deep_rl_deadly_triad,
  title={Deep reinforcement learning and the deadly triad},
  author={Van Hasselt, Hado and Doron, Yotam and Strub, Florian and Hessel, Matteo and Sonnerat, Nicolas and Modayil, Joseph},
  journal={arXiv preprint arXiv:1812.02648},
  year={2018}
}

@inproceedings{kim2019deepmellow_mellowmax_no_target,
  title={Deepmellow: removing the need for a target network in deep q-learning},
  author={Kim, Seungchan and Asadi, Kavosh and Littman, Michael and Konidaris, George},
  booktitle={Proceedings of the twenty eighth international joint conference on artificial intelligence},
  year={2019}
}

@article{ddpg_sac_mobile_robots,
    author = {de Jesus, Junior Costa and Kich, Victor Augusto and Kolling, Alisson Henrique and Grando, Ricardo Bedin and Cuadros, Marco Antonio de Souza Leite and Gamarra, Daniel Fernando Tello},
    title = {Soft Actor-Critic for Navigation of Mobile Robots},
    journal = {Journal of Intelligent and Robotic Systems},
    year = 2021,
    volume={102},
    doi={10.1007/s10846-021-01367-5},
}

@inproceedings{bellemare2017distributional,
    title={A distributional perspective on reinforcement learning},
    author={Bellemare, Marc G and Dabney, Will and Munos, R{\'e}mi},
    booktitle={International conference on machine learning},
    pages={449--458},
    year={2017},
    organization={Pmlr}
}

@inproceedings{schulman2015trust,
  title={Trust region policy optimization},
  author={Schulman, John and Levine, Sergey and Abbeel, Pieter and Jordan, Michael and Moritz, Philipp},
  booktitle={International conference on machine learning},
  pages={1889--1897},
  year={2015},
  organization={Pmlr}
}
